\documentclass{article} 
\usepackage{iclr2025_conference,times}

\usepackage{amsmath,amsfonts,bm}

\def\eqref#1{equation~\ref{#1}}

\def\1{\bm{1}}

\DeclareMathAlphabet{\mathsfit}{\encodingdefault}{\sfdefault}{m}{sl}
\SetMathAlphabet{\mathsfit}{bold}{\encodingdefault}{\sfdefault}{bx}{n}

\usepackage{hyperref}
\usepackage{url}
\usepackage{graphicx}
\usepackage{booktabs} 
\usepackage{enumitem}
\usepackage{array}
\usepackage{multirow}
\usepackage{tcolorbox}
\usepackage{amsmath, amssymb}
\tcbuselibrary{skins, breakable}
\usepackage{tikz}
\usetikzlibrary{decorations.pathmorphing}
\usepackage{xcolor}
\usepackage{algorithm}
\usepackage{algorithmic}
\usepackage{subcaption}
\usepackage{float}

\title{MM-Eureka: Exploring the Frontiers of Multimodal Reasoning with Rule-based Reinforcement Learning}

\author{
\begin{minipage}{\textwidth}
    \vspace{2em}
    \centering
    Fanqing Meng$^{*}$ \quad Lingxiao Du$^{*}$ \quad Zongkai Liu$^{*}$ \quad Zhixiang Zhou$^{*}$ \quad Quanfeng Lu\\[3pt]
    Tiancheng Han \quad Daocheng Fu \quad Botian Shi \quad Wenhai Wang \quad Junjun He \\[3pt]
    Kaipeng Zhang \quad Ping Luo \quad Yu Qiao \quad Qiaosheng Zhang$^{\dagger}$ \quad Wenqi Shao$^{\dagger}$ \\[8pt]
    Shanghai AI Laboratory \quad Shanghai Innovation Institute \\ 
    Shanghai Jiao Tong University \quad The University of Hong Kong
\end{minipage}
}

\iclrfinalcopy 
\begin{document}

\maketitle

\renewcommand{\thefootnote}{\fnsymbol{footnote}}
{\let\thefootnote\relax\footnotetext{
\noindent \hspace{-5mm}$\dagger$ Corresponding Authors: shaowenqi@pjlab.org.cn; zhangqiaosheng@pjlab.org.cn \\
\noindent \hspace{-5mm}\quad \quad $*$ Equal contribution \\
}   }

\begin{abstract}

DeepSeek R1, and o1 have demonstrated powerful reasoning capabilities in the text domain through stable large-scale reinforcement learning. To enable broader applications, some works have attempted to transfer these capabilities to multimodal reasoning. However, these efforts have been limited by the limited difficulty of selected tasks and relatively small training scales, making it challenging to demonstrate strong multimodal reasoning abilities.
To address this gap, we introduce the MMK12 dataset and MM-Eureka with 7B and 32B parameters. The former is a high-quality multimodal mathematics reasoning dataset featuring diverse knowledge domains with human-verified answers and solution processes. The latter is a multimodal model employing rule-based reinforcement learning on MMK12, utilizing online filtering and two-stage training strategy to enhance training stability.
MM-Eureka demonstrates remarkable performance gains in multimodal mathematical reasoning, outperforming previous powerful models like InternVL2.5-78B or InternVL2.5-38B-MPO. In particular, MM-Eureka achieves competitive or superior performance compared to both open-source and closed-source models, and trails slightly behind o1 in multidisciplinary reasoning tasks. 
We open-source our complete pipeline to foster further research in this area. We release all our codes, models, data, etc. at \url{https://github.com/ModalMinds/MM-EUREKA}

\end{abstract}

\begin{figure*}[t]
    \centering
    \includegraphics[width=1.0\linewidth]{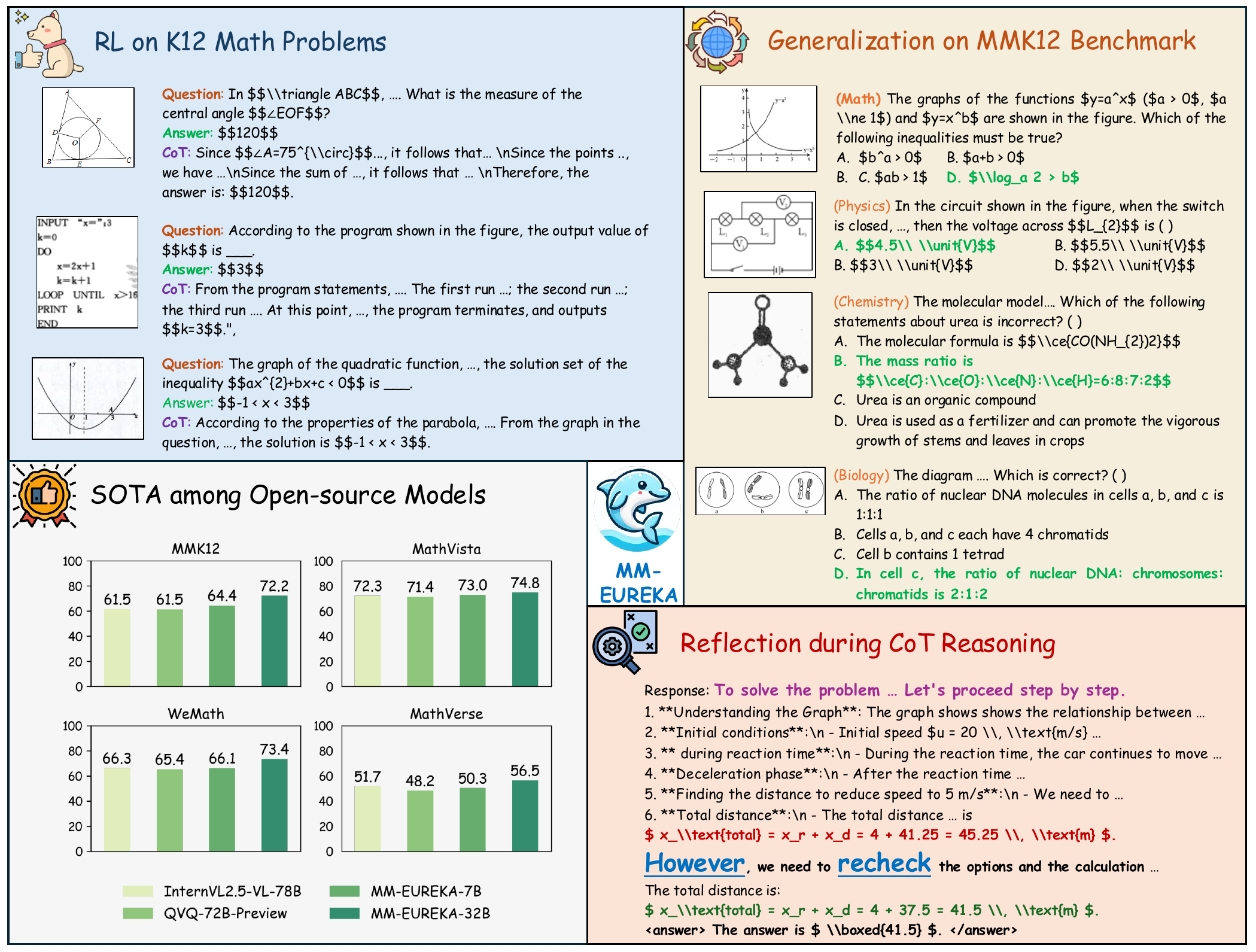}
    \caption{The overview of our proposed MMK12 and MM-Eureka. MMK12 training set has diverse multimodal mathematical questions with verified answer and process, while its evaluation set has multiple-choice questions for each discipline including math, physics, chemistry, and biology. MM-Eureka built on MMK12 has powerful performance in multimodal reasoning and it also exists aha-moment like DeepSeek R1.}
    \label{fig:mmeureka}
\end{figure*}

\section{Introduction}

Large-scale reinforcement learning (RL)~\citep{sutton1998reinforcement} has demonstrated remarkable progress in improving the reasoning ability of Large Language Models (LLMs), particularly in the math and code domains~\citep{openai2024o1, deepseekai2025}. Recent research, such as o1~\citep{openai2024o1} and DeepSeek-R1~\citep{deepseekai2025}, shows that large-scale RL can achieve breakthrough improvements in complex reasoning tasks during post-training phases, sometimes even without supervised fine-tuning (SFT)~\citep{radford2019language}. Despite great success in the text domain, many real-world reasoning tasks such as interpreting scientific diagrams and geometrical reasoning can only be effectively solved with the image input. However, transferring large-scale RL techniques that work well for LLMs to multimodal scenarios remains underexplored.

Recently, many works in the community have attempted to transfer the rule-based RL used in DeepSeek-R1 to multimodal scenarios. However, these works explore relatively small model sizes and fail to achieve stable training over extended periods like DeepSeek-R1. For instance, R1-V \citep{chen2025r1v} demonstrates improvements only in simple counting tasks, which covers limited complexity, while the model size remains modest. LMM-R1 \citep{peng2025lmmr1} achieves gains in accuracy reward for geometric reasoning; however, such success has not been verified in large-scale training with image-text data. Although Kimi k1.5 \citep{kimiteam2025kimik15scalingreinforcement} has achieved competitive results in multimodal reasoning, it has not open-sourced its model or training data to the community. 
Therefore, how to achieve stable rule-based RL training in the multimodal reasoning domain remains an important and unsolved problem for the open-source community.

To address this gap, as shown in Figure \ref{fig:mmeureka}, we first collect MMK12\footnote{K12 refers to the full span of primary and secondary education, and the K12 evaluation sets used in our work primarily contain questions from the secondary level.}, a high-quality and diverse multimodal mathematical reasoning dataset.
problems. 
Subsequently, we implement MM-Eureka with two variants based on the MMK12 training set, i.e. MM-Eureka-7B built on Qwen2.5-VL-Instruct-7B and MM-Eureka-32B built on Qwen2.5-VL-Instruct-32B, which demonstrate excellent performance in multidisciplinary multimodal reasoning. 
For instance, MM-Eureka-7B achieves $66.1$ on MMK12 evaluation sets, only $0.2$ points below Intern2.5-VL-78B. On MathVista, it reaches $73.0$, even surpassing InternVL2.5-VL-78B. MM-Eureka-32B demonstrates stronger performance, scoring $72.3$ on multidisciplinary K12 evaluation sets, which exceeds both Qwen2.5-VL-72B's $70.3$ and closed-source models like Gemini2-Flash, ranking second only to o1's $73.9$. On commonly used multimodal mathematical reasoning benchmarks, MM-Eureka-32B achieves $73.4$ on WeMath, outperforming all open-source models and most closed-source models including Claude3.7 Sonnet. On MathVista, it reaches $74.8$, surpassing all open-source and closed-source models. Both variants demonstrate significant improvements in multidisciplinary K12 and mathematical reasoning performance, outperforming most open-source models of similar sizes.

Specifically, despite growing interest in multimodal reasoning models, high-quality datasets for multimodal mathematical reasoning remain scarce. For example, Geo3k \citep{lu2021inter} and RCOT \citep{deng2024r} focus only on geometry problems. Although MAVIS \citep{zhang2024mavis} comprises data involving geometry and functions, it consists of synthetic data, lacking diversity in images and questions. 
To this end, we construct MMK12, a K12-level multimodal mathematical reasoning dataset. The training set covers a wide range of domains, including functions, geometry, and equations, spanning from elementary through high school curricula. For evaluation, MMK12 includes four disciplines: mathematics, physics, chemistry, and biology. To facilitate testing, we collect $500$ multimodal multiple-choice questions for each discipline and carefully verify that there is no overlap with the training set.

Subsequently, to achieve stable rule-based RL training over extended periods, we utilize the online filter strategy from PRIME \citep{cui2025processreinforcementimplicitrewards} that dynamically filters prompts with zero advantage during training, e.g., those answered either completely correctly or incorrectly. For MM-Eureka-32B, we further introduce a two-stage training strategy: the first stage leverages MMK12 by RL without KL divergence to develop the model's general reasoning abilities while reducing training cost; the second stage fine-tunes the model on Geo3k with KL regularization to mitigate domain-specific deficiencies and enhance training stability. We use the GRPO \citep{deepseekai2025} as our basic RL approach, which offers higher efficiency compared to the commonly used PPO \citep{schulman2017proximal}. These strategies enable us to achieve stable, long-horizon RL training and consistent performance gains for both 7B and 32B model variants, while maintaining high training efficiency.

Through the journey of developing MM-Eureka, we have several findings.
\textbf{First,} it is difficult for the model to acquire new knowledge through RL training. Instead, performance improvements come from increasing the probability that the model generates correct answers during inference.
\textbf{Second,} we discover that simple rule-based RL exhibits outstanding generalization capabilities. Training solely on mathematical data leads to simultaneous improvements in physics, chemistry, biology, and other disciplines.
\textbf{Third,} rule-based RL generalizes better than other post-training strategies such as SFT \citep{ouyang2022traininglanguagemodelsfollow} and COT SFT \citep{guo2024mammothvlelicitingmultimodalreasoning} across various tasks.

Our goal is to share our implementation experiences and complete the open-source pipeline with the community, including data, code, and models. We believe this comprehensive open-source framework would help the community better explore the multimodal reasoning task. The main contributions are summarized as follows:

\begin{itemize}[leftmargin=10pt]
\item We utilize an online filter strategy and introduce a two-stage training strategy to address the collapse issues encountered during RL training of large-scale VLMs, achieving stable rule-based RL training for large-scale VLMs.
\item We present MM-Eureka-7B and MM-Eureka-32B. Extensive experimental results on various downstream tasks demonstrate that they are top performers among open-source models in the multimodal reasoning domain. For example, MM-Eureka-32B scores only $1.7$ points below o1 on the multidisciplinary evaluation set of MMK12; MM-Eureka-7B achieves $73.0$ on MathVista \citep{lu2024mathvistaevaluatingmathematicalreasoning}, surpassing InternVL2.5-78B \citep{chen2024expanding}.
\item We open-source all our models, code, and collected high-quality multimodal mathematical reasoning data. Compared to existing open-source repositories, we support a wider range of RL algorithms and include much higher-quality data.

\end{itemize}

\section{Related Work}

\subsection{Language Reasoning Model}

LLMs have demonstrated impressive performance across a wide range of tasks, yet more complex challenges require these models to exhibit human-like reasoning capabilities. As a result, enhancing the reasoning ability of LLMs has become a critical research focus. Reinforcement Learning from Human Feedback (RLHF), particularly Proximal Policy Optimization (PPO)~\citep{PPO}, has shown promise in enabling LLMs to learn reasoning abilities effectively. However, the PPO training process is computationally intensive and complex, prompting the development of simplified alternatives such as Direct Preference Optimization (DPO)~\citep{rafailov2023direct}. While DPO alleviates some training difficulties, its reliance on offline data can limit model performance. To address these limitations, methods like Group Relative Policy Optimization (GRPO)~\citep{deepseekai2025}, REINFORCE Leave-One-Out (RLOO)~\citep{RLOOKoolHW19a, RLOOAhmadianCGFKPUH24}, and Reinforce++~\citep{hu2025reinforce++} have been introduced. Notably, Deepseek R1~\citep{deepseekai2025} reveals that pure RL can encourage LLMs to actively engage in reasoning, including self-reflection and error correction. Despite these advancements, research on improving the reasoning capabilities of multimodal large models remains relatively scarce, highlighting an important direction for future exploration.

\subsection{Vision-Language Reasoning Model}

Currently, the leading models in multimodal reasoning are closed-source systems such as GPT-4o~\citep{hurst2024gpt} and Kimi-VL~\citep{kimiteam2025kimivltechnicalreport}. In contrast, the open-source community remains noticeably behind, still in the early stages of exploration. Recent concurrent efforts have begun to explore the use of RL to enhance the visual reasoning capabilities of vision-language reasoning models (VLMs), aiming to trigger an ``Aha Moment" in visual reasoning. LMM-R1~\citep{peng2025lmmr1} strengthens visual reasoning through a two-stage rule-based RL approach; however, its primary reasoning performance gains are derived from text-only datasets rather than genuinely multimodal datasets. R1-V~\citep{chen2025r1v} investigates rule-based RL within specific subdomains, such as geometric reasoning and object counting tasks, but falls short of addressing more complex reasoning challenges. Reason-RFT~\citep{tan2025reasonrftreinforcementfinetuningvisual}, on the other hand, relies on SFT with COT reasoning activation data to achieve an effective cold start before the RL training phase. 
In this paper, our objective is to develop an effective, stable, and comprehensive open-source training pipeline for multimodal reasoning models, including datasets, code, and models. Our work aims to advance the growth and innovation of the open-source community.

\section{MMK12: Multimodal Mathematic K12-Level Dataset} \label{sec:dataset}

\begin{table}[htbp]
  \centering
  \scalebox{0.8}{
  \begin{tabular}{lccccc}
    \toprule
    & \textbf{Scope} & \textbf{Type} & \textbf{Img. Source} & \textbf{QA Source} & \textbf{CoT Answer Source} \\
    \midrule
    MAVIS \citep{zhang2024mavis} & Geo \& Func  & MCQ \& FB & Synthetic & Synthetic Engine & GPT-4o \\
    Geo3k \citep{lu2021inter} & Geo & FB & Real world & Real world  &  None\\
    RCOT \citep{deng2024r} & Geo & MCQ \& FB & Synthetic & Synthetic Engine & GPT-4o \\
    MultiMath \citep{peng2024multimath} &  Diverse & MCQ \& FB & Real World & GPT-4o & GPT-4o \\
    \midrule
    MMK12 & Diverse & FB & Real World  &  Real World  &   Real World\\
    \bottomrule
  \end{tabular}
  }
  \caption{Comparison of dataset characteristics with other multimodal mathematical reasoning datasets. MMK12 comprises more diverse and high quality questions, with guaranteed correct answers and solution processes.}
  \label{tab:dataset_comparison}
\end{table}

As shown in Table \ref{tab:dataset_comparison}, current multimodal mathematical reasoning datasets have limited scope and face challenges in ensuring answer correctness. For instance, while RCOT and MAVIS maintain answer accuracy through synthetic engine-generated QA pairs, this approach restricts problem diversity. Geo3k manually collected 3,000 geometry problems with verified answers, but it focuses solely on geometry examples. Although MultiMath gathers problems from real-world scenarios to ensure diversity, its reference answers generated by GPT-4o cannot guarantee correctness. 

To address these limitations, we introduce MMK12, a new dataset comprising over 15,000 multimodal mathematical reasoning problems across a wide range of domains, including geometry, functions, and graphical reasoning. Each problem is accompanied by a standard reference answer and a detailed step-by-step solution to ensure both accuracy and interpretability.

\begin{figure*}[t]
    \centering
    \includegraphics[width=1.0\linewidth]{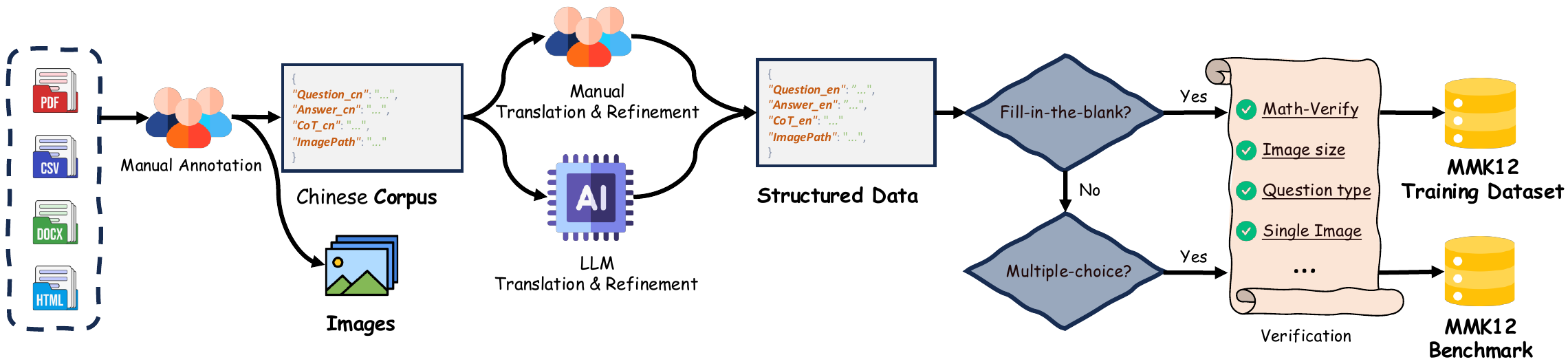}
    \caption{The construction overview of MMK12. We collect diverse K12-level multimodal math problems from multiple sources, convert them to standardized English using LLMs, and verify all content for accuracy. The resulting MMK12 dataset includes a training set of 15,616 samples and a test set with 500 multiple-choice questions each for math, physics, chemistry, and biology.}
    \label{fig:mmk12_construct}
\end{figure*}

In the following, we describe the construction and cleaning process of MMK12. As illustrated in Figure \ref{fig:mmk12_construct}, we first collect a diverse set of multimodal mathematics problems and corresponding answers from a variety of Chinese mathematics textbooks and examination papers, covering grades from elementary to high school. These questions, answers, and COT processes are subsequently translated and refined into English, with the help of LLM. To minimize false positives during training, we retain only fill-in-the-blank problems. Besides, we use Math-Verify\footnote{https://github.com/huggingface/Math-Verify} to parse the answer, ensuring data reliability for RL training.
For the MMK12 evaluation set, we follow a similar construction procedure, but exclusively select multiple-choice questions to facilitate reliable and efficient evaluation.

In total, the MMK12 training dataset comprises 15,616 multimodal fill-in-the-blank mathematics problems, including 455 from elementary school, 9,776 from middle school, and 5,385 from high school. Each sample includes the question, an image, the final answer, and a CoT-formatted solution process.
The MMK12 evaluation set has 2,000 multimodal multiple-choice questions, including math, physics, chemistry, and biology. Each item includes a question, a corresponding image, and the correct answer choice. Some examples in MMK12 are shown in Figure \ref{fig:mmeureka}.

MMK12 encompasses mathematical problems across various knowledge domains, including geometry, functions, spatial reasoning, and more. Some important categories are introduced as follows:

\begin{itemize}[leftmargin=10pt]

\item Function Reasoning: This task requires models to understand function concepts, analyze function graphs and expressions, and apply function properties to solve problems. This type of reasoning develops the model's ability to comprehend abstract mathematical concepts, fostering its capability to identify function characteristics, determine critical points, and analyze function behavior.

\item Geometric Reasoning: This task involves applying spatial relationships, geometric theorems, and properties of shapes. Through geometric reasoning training, models enhance their spatial visualization, logical deduction, and formalization abilities for geometry problems, enabling them to solve complex problems in both plane and solid geometry.

\item Pattern Reasoning: This type of task focuses on understanding flow diagrams and recognizing patterns in visual sequences. Models need to discover patterns, predict rule-based changes, or understand logical relationships in visual content. This task examines the model's pattern recognition abilities, inductive reasoning skills, and visual logical thinking.

\end{itemize}

Benefiting from a standardized data construction process with substantial human involvement and verification, MMK12 ensures both diversity in questions and correctness in answers and solution processes, making it suitable for various training methods such as RL and SFT for multimodal mathematical reasoning models. The MMK12 evaluation set also provides a convenient and accurate means to test models' multidisciplinary reasoning capabilities.

\section{Method}

\subsection{Basic Settings}

We employ Qwen2.5-VL~\citep{bai2023qwenvlversatilevisionlanguagemodel} with 7B and 32B as the base models. Our RL algorithm is similar to DeepSeek-R1~\citep{deepseekai2025}, using rule-based format rewards $r_\text{format}\in\{0, 0.5\}$, accuracy rewards $r_\text{accuracy}\in\{0, 1\}$ as reward function and GRPO as base RL algorithm for training. Furthermore, we develop a multimodal input RL framework based on OpenRLHF \citep{hu2024openrlhf}, which is compatible with commonly used models such as InternVL~\citep{chen2024expanding} and QwenVL, supporting various model sizes and RL algorithms. In the following sections, we provide detailed settings for our RL training.

\subsection{Reward Function}

Following DeepSeek-R1, we also adopt the simple rule-based reward function rather than using outcome or process reward models, thereby alleviating reward hacking~\citep{gao2022scalinglawsrewardmodel}. Specifically, we use two types of rewards: accuracy reward and format reward. The former uses Math-Verify to extract the answer from model responses and compare it with the reference one, returning 1 or 0 based on correctness; the latter checks whether the response follows the specified format ($\texttt{<think>}...\texttt{</think>}\texttt{<answer>}...\texttt{</answer>}$), returning 0.5 or 0 based on compliance. We find that this simple and sparse reward is sufficient to significantly improve the model's multimodal reasoning ability.

\subsection{Advantage Estimation and Policy Update}

Group Relative Policy Optimization (GRPO)~\citep{deepseekai2025} is a widely adopted RL algorithm that eliminates the need for training a complex critic model by leveraging intra-group relative performance to optimize the policy model. 
Specifically, for each query $\bold{x}$, the model generates a group of $G$ responses $\{\bold{y}^{(1)},\bold{y}^{(2)},\cdots,\bold{y}^{(G)}\}$. Subsequently, for each query with $G$ responses, GRPO computes the relative advantage of each response based on their rewards, which are determined by a reward model, as follows:
\begin{align*}
A^{(i)} = 
 \frac{r^{(i)} - \text{mean}(\{r^{(j)}\}_{j=1}^{G})}{\text{std}(\{r^{(j)}\}_{j=1}^{G})}, \quad i=1,\cdots,G.
\end{align*}
Using the computed advantages, GRPO then optimizes the policy via the PPO-clip loss augmented with a directly imposed KL penalty term:
\begin{align}\label{eq: grpo}
J_\text{PPO}(\theta) &= -\mathbb{E}_{\bold{x}\sim \mathcal{D}, \{\bold{y}^{(i)}\}_{i=1}^G\sim \pi_{\theta}(\cdot|\bold{x})}
\\\Bigg[ \frac{1}{G}&\sum_{i=1}^G\frac{1}{|\bold{y}^{(i)}|}\sum_{t=1}^{|\bold{y}^{(i)}|}
\Bigg(\min\Big( r^{i,t}(\theta)A^{(i)},\text{clip}\big(r^{i,t}(\theta), 1-\epsilon, 1+\epsilon\big) A^{(i)}\Big)
-\beta D_{\text{KL}}^{i, t}(\pi_\theta, \pi_\text{ref})
\Bigg)\Bigg],
\nonumber
\end{align}
where
\begin{align*}
    r^{i,t}(\theta) = \frac{\pi_{\theta}(y_t^{(i)}|\bold{x}, \bold{y}^{(i)}_{<t})}{\pi_{\theta_\text{old}}(y_t^{(i)}|\bold{x}, \bold{y}^{(i)}_{<t})} 
    \text{ and }
    D_\text{KL}^{i,t}(\pi_\theta, \pi_\text{ref})=
    \frac{\pi_\text{ref}(y_t^{(i)}|\bold{x}, \bold{y}^{(i)}_{<t})}{\pi_{\theta}(y_t^{(i)}|\bold{x}, \bold{y}^{(i)}_{<t})} - 1 - \log \frac{\pi_\text{ref}(y_t^{(i)}|\bold{x}, \bold{y}^{(i)}_{<t})}{\pi_{\theta}(y_t^{(i)}|\bold{x}, \bold{y}^{(i)}_{<t})}.
\end{align*}
In addition, to mitigate loss spikes during training caused by excessively large policy ratios combined with negative advantages, we further constrain the policy ratio within $[0, c]$ beforehand, i.e., we replace $r^{i,t}(\theta)$ with $\text{clip}(r^{i,t}(\theta), 0, c)$ in Eq.(\ref{eq: grpo}). In practice, we set the default value of $c$ to 3.

\subsection{Online Filtering}

To further enhance training stability, we adopt the online prompt filtering strategy proposed in PRIME~\citep{cui2025processreinforcementimplicitrewards}.
To ensure sufficient gradient information throughout RL training, we filter out prompts with responses that are either completely correct or completely incorrect during training, as their corresponding advantages under GRPO are zero. The detailed process is shown in Algorithm~\ref{alg: online filter}.

\begin{figure*}[t]\label{fig: online filter}
    \centering
    \includegraphics[width=0.99\linewidth]{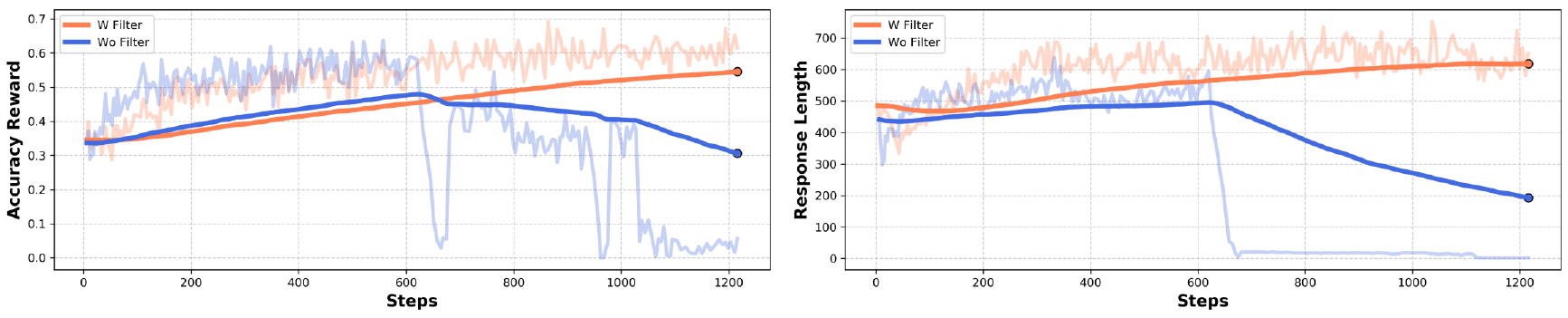}
    \caption{Comparison of the use of online filtering. Using online filter makes the RL training more stable in the long term, with consistently increasing accuracy reward and response length.}
    \label{fig:nofilter-log}
\end{figure*}

We present an ablation study on the online filtering mechanism in Figure~\ref{fig:nofilter-log}. Models trained with the online filter strategy maintain relatively stable trends in both the accuracy-based reward and response length throughout training. In contrast, models without the online filter show an initial improvement in accuracy, but soon experience a sharp decline, with accuracy eventually approaching zero and response lengths significantly shortening in the later stages. These findings suggest that the online filter plays an important role in stabilizing RL training and can help prevent model collapse during optimization.

\begin{algorithm*}[ht!]
\caption{Online Filter}
\label{alg: online filter}
\textbf{Input}: Prompt dataset $\mathcal{D}$; initial policy $\boldsymbol{\pi}_\theta$; reward model $\boldsymbol{R}$; buffer size $N_B$; hyperparameters $K_1, K_2, G, \epsilon_\text{acc}^\text{lower}, \epsilon_\text{acc}^\text{upper}$\\
\textbf{Output}: $\boldsymbol{\pi}_{\theta}$
\begin{algorithmic}[1] 
\STATE Buffer $\mathcal{B}\gets \{\}$
\FOR{$\text{iteration}=1,\cdots, K_1$}
    \STATE Sample a batch of prompts $\mathcal{Q}\sim\mathcal{D}$
    \FOR{each prompt $\mathbf{x}\in \mathcal{Q}$}
        \STATE Generate $G$ responses $\{\mathbf{y}^{(i)}\}_{i=1}^{G} \sim \boldsymbol{\pi}_{\theta}(\cdot|\mathbf{x})$
        \STATE Compute the reward $r^{(i)} = \boldsymbol{R}(\mathbf{x}, \mathbf{y}^{(i)})$ for $i=1,\cdots,G$
        \STATE Compute the accuracy $\mathcal{C}_{\mathbf{x}} = |\{\mathbf{y}^{(i)}|r^{(i)}=1\}|$
        \IF{$\epsilon_\text{acc}^\text{lower} \le \mathcal{C}_{\mathbf{x}} \le \epsilon_\text{acc}^\text{upper}$}
        \STATE Add the sample to the buffer $\mathcal{B}\gets\mathcal{B}\cup \{(\mathbf{x}, \mathbf{y}^{(i)}, r^{(i)})\}_{i=1}^{G} $
        \ENDIF
    \ENDFOR
    \IF{$|\mathcal{B}| \ge N_B$}
        \FOR{$\text{Epoch}=1,\cdots, K_2$}
            \STATE  Update policy $\boldsymbol{\pi}_\theta$ on $\mathcal{B}[0{:}N_B]$ by any RL algorithm
        \ENDFOR
        \STATE Buffer $\mathcal{B}\gets \{\}$
    \ENDIF
\ENDFOR
\end{algorithmic}
\end{algorithm*}

\subsection{Two-stage Training} \label{sec:twostage}

During RL training of MM-Eureka-32B, we observe that maintaining training stability becomes increasingly challenging as model size scales up. Specifically, we find that models tend to experience sudden collapse during training, characterized by an abrupt drop of the accuracy reward to near-zero values, as shown in Figure \ref{fig:twostage-log} (a). After analysis, we find that these collapses are usually preceded by sharp increases in the policy ratio (indicating excessively large updates from old to new policies) like Figure \ref{fig:twostage-log} (c). Additionally, unlike the 7B model which shows comprehensive improvement when trained solely on K12 data, the 32B model trained only on MMK12 exhibits performance degradation in specific domains such as geometry.

To address these challenges, we propose a two-stage training strategy for large models. In the first stage, we apply the GRPO algorithm without KL regularization, using MMK12 data collected as detailed in Section~\ref{sec:dataset} to broadly enhance the model's general reasoning capabilities. In the second stage, we incorporate a KL divergence into GRPO to constrain update magnitudes and improve training stability. Concurrently, we augment the training set with domain-specific Geo3k data to address identified performance gaps. As shown in Section~\ref{sec:twostageabli}, this two-stage approach enables stable training and yields further performance gains across nearly all evaluated benchmarks.

\section{Experiments}

We present our experimental setup in Section \ref{sec:exp-setup}, where we indicate the training details. After that, we provide an overview of the selected baselines and evaluation benchmarks in Section \ref{sec:exp-baseline} and Section \ref{sec:exp-eval}, we demonstrate the superior effectiveness of MM-Eureka-7B and MM-Eureka-32B through extensive experiments in Section \ref{sec:exp-res}. 

\subsection{Experiments Setup}\label{sec:exp-setup}

We design our prompt template following the format used in DeepSeek-R1, wherein the system prompt explicitly specifies the required output structure, including the use of \texttt{<answer>} tags to separate the reasoning process from the final answer. Detailed prompt configurations are provided in Table~\ref{tab:promt-eureka}.

For training hyperparameters, both the rollout batch size and training batch size are set to $128$, with $8$ rollouts generated per sample. Sampling is conducted with a temperature of $1.0$ to encourage response diversity, and optimization is performed using a learning rate of $1{\times}10^{-6}$. 
Besides, for MM-Eureka-32B, we adopt a two-stage training scheme. In Stage 1, KL divergence is disabled to promote exploration and policy flexibility during the early phase. In Stage 2, a small KL penalty  $1{\times}10^{-3}$ is introduced to stabilize training.


\begin{table}[h]
    \centering
    \caption{Prompt setting for MM-Eureka.}
    \label{tab:promt-eureka}
    \scalebox{1.0}{
    \begin{tabular}{p{12cm}}
        \toprule
        \textbf{SYSTEM:} Solve the question. The user asks a question, and you solve it. You first think about the reasoning process in the mind and then provide the user with the answer. The answer is in latex format and wrapped in \$...\$. The final answer must be wrapped using the \textbackslash boxed\{\} command. Th answer should be enclosed within \textless answer\textgreater \textless /answer\textgreater tags, i.e., Since \$1+1=2\$, so the answer is \$2\$. \textless answer\textgreater The answer is \$\textbackslash boxed\{2\}\$ \textless /answer\textgreater, which means the final answer assistant's output should start with \textless answer\textgreater and end with \textless /answer\textgreater. \\  
        \textbf{USER:} \textless image\textgreater \textcolor{red}{\{\{question\}\}} \\
        \bottomrule
    \end{tabular}
    }
\end{table}

\subsection{Baselines}\label{sec:exp-baseline}

To comprehensively evaluate the effectiveness of MM-Eureka, we compare it against a diverse set of baselines, including both closed-source and open-source systems.

\paragraph{Closed-Source Models.}
We include several leading proprietary models in our comparison. GPT-4o \citep{hurst2024gpt}, Claude3.7-Sonnet \citep{claude3.7sonnet}, and Gemini2-flash \citep{team2023gemini} are general-purpose multimodal models without dedicated reasoning optimization, while o1 \citep{openai2024o1} is explicitly designed to enhance reasoning capabilities. These models serve as strong references for the current state of the art.

\paragraph{Open-Source General Models.}
This group comprises high-capacity vision-language models trained for general purposes, including Qwen-2.5-VL \citep{bai2023qwenvlversatilevisionlanguagemodel} and InternVL2.5 \citep{chen2024expanding} across various model sizes (7B to 78B). These models are primarily pretrained or instruction-tuned on large-scale image-text datasets. They offer a baseline for evaluating the impact of reasoning-specific post-training.

\paragraph{Open-Source Reasoning Models.}
We further include open-source models explicitly fine-tuned for reasoning, such as InternVL2.5-MPO variants \citep{wang2024enhancingreasoningabilitymultimodal} , QVQ-72B-Preview \citep{qvq-72b-preview} , Adora-7B \citep{gui2025adora} , R1-Onevision \citep{yang2025r1onevision} , and OpenVLThinker \citep{deng2025openvlthinker}. These models employ various strategies, including SFT \citep{ouyang2022traininglanguagemodelsfollow}, DPO \citep{rafailov2023direct}, and rule-based reinforcement learning. They represent the most competitive open-source efforts in multimodal reasoning.

\subsection{Benchmarks}\label{sec:exp-eval}

To assess the performance of our model, we conduct evaluations on multiple benchmark datasets, including MathVista(testmini)~\citep{lu2024mathvistaevaluatingmathematicalreasoning}, MathVerse(testmini)~\citep{zhang2024mathversedoesmultimodalllm}, MathVision(test)~\citep{wang2024measuringmultimodalmathematicalreasoning}, OlympiadBench(EN-OE split)~\citep{he2024olympiadbenchchallengingbenchmarkpromoting} and WeMath \citep{qiao2024we}. MathVista is one of the most widely used multimodal mathematical benchmarks, offering a diverse set of problems that span general visual question answering, figure question answering, logic, algebra, and geometry. MathVerse, on the other hand, focuses specifically on the model’s ability to comprehend images, with tasks categorized into areas such as algebra and geometry. MathVision takes this a step further by emphasizing more abstract visual understanding, testing the model’s capacity for recognizing and reasoning beyond conventional mathematical contexts. OlympiadBench presents graduate-level mathematical competition problems, from which we select the English questions for evaluation. WeMath complements these by providing fine-grained diagnostic insights into model reasoning behavior, using a large-scale, hierarchically annotated problem set to assess knowledge mastery and generalization.

Beyond these established benchmarks, both our model and the baselines are further evaluated on the proposed MMK12 Benchmark, as introduced in Section~\ref{sec:dataset}. Unlike OlympiadBench, which targets advanced graduate-level mathematical reasoning, MMK12 focuses on assessing the model’s ability to solve fundamental multimodal multidisciplinary problems commonly encountered in K12 education. MMK12 enables a comprehensive evaluation of a model's ability across disciplines beyond mathematics.

During evaluation, we adopt greedy decoding with a temperature of $0$, ensuring deterministic outputs. We do not use beam search, and both top-$p$ and top-$k$ sampling are disabled.

\subsection{Quantitative Results}\label{sec:exp-res}

In this section, we demonstrate the superiority of MM-Eureka through its performance on the benchmarks introduced in Section \ref{sec:exp-eval}. For the multimodal mathematical evaluation sets, we primarily use official results or results provided in VLMEvalKit \citep{duan2024vlmevalkit} for baseline comparisons. When such results are unavailable, we conduct evaluations using the testing methodologies provided in VLMEvalKit within our vllm-based \citep{kwon2023efficient} inference framework. Specifically, for o1, due to resource constraints, we randomly sample only 500 instances for testing. For the MMK12 evaluation set, considering it consists entirely of multiple-choice questions, we use the same Chain-of-Thought prompt structure from WeMath \citep{qiao2024we} to test models within our inference framework. Because WeMath also consists entirely of multiple-choice questions. Results evaluated by ourselves are presented in italics.

\paragraph{Mathematics}

\begin{table}[h]
    \centering
    \caption{Performance comparison across different multimodal mathematical benchmarks. Bold indicates the top performer with the open-source models, while underline indicates the second best performer within open-source models. For some official results that are relatively different from our evaluation results, we mark our test results after '/' .}\vspace{5pt}
    \scalebox{0.93}{
    \begin{tabular}{lccccc}
        \toprule
        \textbf{Model} & \textbf{MathVista } & \textbf{MathVerse } & \textbf{Mathvision } & \textbf{OlypamidBench} & \textbf{WeMath} \\
        \midrule
        \multicolumn{6}{l}{\textbf{Closed-Source Models}} \\
        Claude3.7-Sonnet & 66.8 & \textit{52.0} & 41.3 & \textit{48.9} & 72.6   \\
        GPT-4o & 63.8 & 50.2 & 30.4 & \textit{35.0} & 68.8  \\
        o1 & 73.9 & 57.0 & 60.3 & \textit{68.0}* & \textit{98.7}*   \\
        Gemini2-flash & 70.4 & \textit{59.3} & 41.3 & \textit{51.0} & 71.4 \\
        \midrule
        \multicolumn{6}{l}{\textbf{Open-Source General Models}} \\
        InternVL2.5-VL-8B & 64.4 & 39.5 & 19.7 & \textit{12.3} & 53.5   \\
        Qwen-2.5-VL-7B & 68.2 & 47.9 & 25.4 & \textit{20.2} & 62.1  \\
        InternVL2.5-VL-38B & 71.9 & 49.4 & 31.8 & \textit{32.0} & 67.5  \\
        Qwen-2.5-VL-32B & \underline{74.7}/\textit{71.7} & \textit{49.9} & \textbf{40.1} & \textit{30.0} & \textit{69.1}    \\
        InternVL2.5-VL-78B & 72.3 & 51.7 & 32.2 & \textit{31.1} & 66.3   \\
        Qwen-2.5-VL-72B & \textbf{74.8} & \textbf{57.6} & \underline{38.1} & \textbf{\textit{40.4}} & \underline{72.4}   \\
        \midrule
        \multicolumn{6}{l}{\textbf{Open-Source Reasoning Models}} \\
        InternVL2.5-8B-MPO & 68.9 & \textit{35.5} & \textit{21.5} & \textit{7.8} & 53.5 \\
        InternVL2.5-38B-MPO & 73.8 & \textit{46.5} & \textit{32.3} & \textit{25.6} & 66.2 \\
        QVQ-72B-Preview & 71.4 & 48.2 & 35.9 & \textit{33.2} & 65.4 \\
        Adora-7B & 73.5 & \textit{50.1} & \textit{23.0} & \textit{20.1} & \textit{64.2}   \\
        R1-Onevision-7B & 64.1 & 47.1 & 29.9/\textit{23.5} & \textit{17.3} & 61.8   \\
        OpenVLThinker-7B & 70.2 & 47.9 & 25.3 & \textit{20.1} & \textit{64.3}  \\
        \midrule
        \multicolumn{6}{l}{\textbf{Ours}} \\
        MM-Eureka-7B & 73.0 & 50.3 & 26.9 & 20.1 & 66.1  \\
        MM-Eureka-32B & \textbf{74.8} & \underline{56.5} & 34.4 & \underline{35.9} & \textbf{73.4}  \\
        \bottomrule
    \end{tabular}
    }
    \label{tab:benchmark_comparison}
\end{table}

\footnotetext{Results of o1 based on random sampling of 500 test instances due to resource constraints.}
\footnotetext{Results of R1-Onevision in Mathvision is 23.5 from our evaluation, which is far behind its official report as 29.9.}

In Table \ref{tab:benchmark_comparison}, we present a comprehensive summary of MM-Eurekas' performance across different multimodal mathematical reasoning tasks. It demonstrates that both MM-Eureka-7B and MM-Eureka-32B consistently outperform similar-sized open-source baselines across almost all tasks. In particular, MM-Eureka-7B achieves 73.0 on Mathvista, surpassing InternVL-78B by 0.7\% and the reasoning-focused InternVL2.5-38B-MPO by 0.2\%. On WeMath, it similarly approaches the performance of InternVL-78B. MM-Eureka-32B exceeds all current open-source models of comparable size on all benchmarks except MathVision, with average performance approaching Qwen-72B-Instruct. It even outperforms the closed-source model Claude3.7 Sonnet on WeMath. In summary, MM-Eureka establishes itself as the top performer among same-sized open-source models across almost all multimodal mathematical reasoning tasks, regardless of whether considering the 7B or 32B version.

However, when compared to closed-source models, MM-Eureka-32B still shows considerable performance gaps. While it outperforms o1 on Mathvista, it lags significantly behind closed-source multimodal reasoning models like o1 on more challenging benchmarks such as Mathvision or OlympiadBench.

\paragraph{MMK12}

\begin{table}[h]
    \centering
    \caption{Performance comparison across different disciplines in MMK12. Bold indicates the top performer with the open-source models, while underline indicates the second best performer within open-source models.}\vspace{5pt}
    \scalebox{1.0}{
    \begin{tabular}{lccccc}
        \toprule
        \textbf{Model} & \textbf{Mathematics} & \textbf{Physics} & \textbf{Chemistry} & \textbf{Biology} & \textbf{Avg.} \\
        \midrule
        \multicolumn{6}{l}{\textbf{Closed-Source Models}} \\
        Claude3.7-Sonnet & 57.4 & 53.4 & 55.4 & 55.0 & 55.3 \\
        GPT-4o & 55.8 & 41.2 & 47.0 & 55.4 & 49.9 \\
        o1 & 81.6 & 68.8 & 71.4 & 74.0 & 73.9 \\
        Gemini2-flash & 76.8 & 53.6 & 64.6 & 66.0 & 65.2 \\
        \midrule
        \multicolumn{6}{l}{\textbf{Open-Source General Models}} \\
        InternVL2.5-VL-8B & 46.8 & 35.0 & 50.0 & 50.8 & 45.6 \\
        Qwen-2.5-VL-7B & 58.4 & 45.4 & 56.4 & 54.0 & 53.6 \\
        InternVL2.5-VL-38B & 61.6 & 49.8 & 60.4 & 60.0 & 58.0 \\
        Qwen-2.5-VL-32B & 71.6 & 59.4 & 69.6 & 66.6 & 66.8 \\
        InternVL2.5-VL-78B & 59.8 & 53.2 & 68.0 & 65.2 & 61.6 \\
        Qwen-2.5-VL-72B & \textbf{75.6} & \textbf{64.8} & 69.6 & 72.0 & 70.5 \\
        \midrule
        \multicolumn{6}{l}{\textbf{Open-Source Reasoning Models}} \\
        InternVL2.5-8B-MPO & 26.6 & 25.0 & 42.4 & 44.0 & 34.5 \\
        InternVL2.5-38B-MPO & 41.4 & 42.8 & 55.8 & 53.2 & 48.3 \\
        QVQ-72B-Preview & 61.4 & 57.4 & 62.6 & 64.4 & 61.5 \\
        Adora & 63.6 & 50.6 & 59.0 & 59.0 & 58.1 \\
        R1-Onevision & 44.8 & 33.8 & 39.8 & 40.8 & 39.8 \\
        OpenVLThinker-7 & 63.0 & 53.8 & 60.6 & 65.0 & 60.6 \\
        \midrule
        \multicolumn{6}{l}{\textbf{Ours}} \\
        MM-Eureka-7B & 71.2 & 56.2 & 65.2 & 65.2 & 64.5 \\
        MM-Eureka-32B & \underline{74.6} & \underline{62.0} & \textbf{75.4} & \textbf{76.8}& \textbf{72.2} \\
        \bottomrule
    \end{tabular}
    }
    \label{tab:subject_comparison}
\end{table}

Beyond validating our model's superiority on widely-used multimodal mathematical reasoning benchmarks like MathVista, it is necessary to test its capabilities and generalization across multidisciplinary reasoning domains using questions absent from the training set (e.g., physics, chemistry, and biology). For this purpose, we employ the MMK12 dataset constructed in Section~\ref{sec:dataset}, which can effectively measure models' multimodal reasoning capabilities across multiple disciplines.

As shown in Table \ref{tab:subject_comparison}, MM-Eureka-32B demonstrates multidisciplinary capabilities only marginally behind o1 by 1.7\%, while outperforming larger-scale models such as Qwen-2.5-VL-72B and Gemini2-Flash-Thinking. MM-Eureka-7B also surpasses several similarly-sized multimodal reasoning models, including OpenVLThinker-7B, with overall performance exceeding InternVL2.5-VL-78B and only slightly behind Qwen-2.5-VL-32B.
Additionally, we observe several interesting findings: 1) Despite being trained exclusively on fill-in-the-blank questions, our models maintain strong instruction-following abilities for multiple-choice questions with improved performance. 2) Even with training solely on mathematics problems, the models exhibit enhanced capabilities in physics, chemistry, and biology. Specifically, MM-Eureka-7B shows improvements of 9.8\% and 11.2\% in chemistry and biology, respectively, demonstrating the remarkable generalization capacity of our straightforward RL strategy.

\subsection{Qualitative Results}\label{sec:exp-qualres}

Appendix \ref{appendix:qualitative} illustrates representative examples comparing MM-Eureka-32B with its base model Qwen2.5-VL-32B-Instruct across four subjects: mathematics, physics, chemistry, and biology. These qualitative cases reveal that our model, after reinforcement learning, demonstrates significantly enhanced reasoning capabilities. Specifically, MM-Eureka-32B is able to better apply known concepts and perform multi-step deduction to arrive at the correct answers. In contrast, the base model often shows surface-level understanding---it may recall relevant facts but fails to apply them coherently in the problem-solving context.

For instance, in the physics example (Figure \ref{fig:qualitative_phy}), the question involves analyzing the instantaneous contact force between two objects placed on a vertical spring. While both models recognize key physical quantities such as mass, weight, and spring force, only MM-Eureka-32B correctly applies Newton’s second law to derive the acceleration of the system and compute the contact force. It identifies that the net external force due to the added mass leads to a downward acceleration, and accurately calculates the resulting contact force between object A and B as $24N$. In contrast, the base model Qwen2.5-VL-32B-Instruct incorrectly concludes that the contact force equals the full weight of object B (i.e., $40N$), failing to consider the transient acceleration of the system and misapplying Newtonian mechanics. 

These examples highlight MM-Eureka’s improved ability to perform reasoning and its potential to generalize reasoning patterns across disciplines. Further cases in mathematics, chemistry, and biology show similar trends, where our model more effectively breaks down problems into intermediate steps and synthesizes knowledge to reach accurate conclusions.

\section{Discussion}

\subsection{Are knowledge and reasoning decoupled?}

\begin{figure}[ht]
    \centering
    \includegraphics[width=0.9\linewidth]{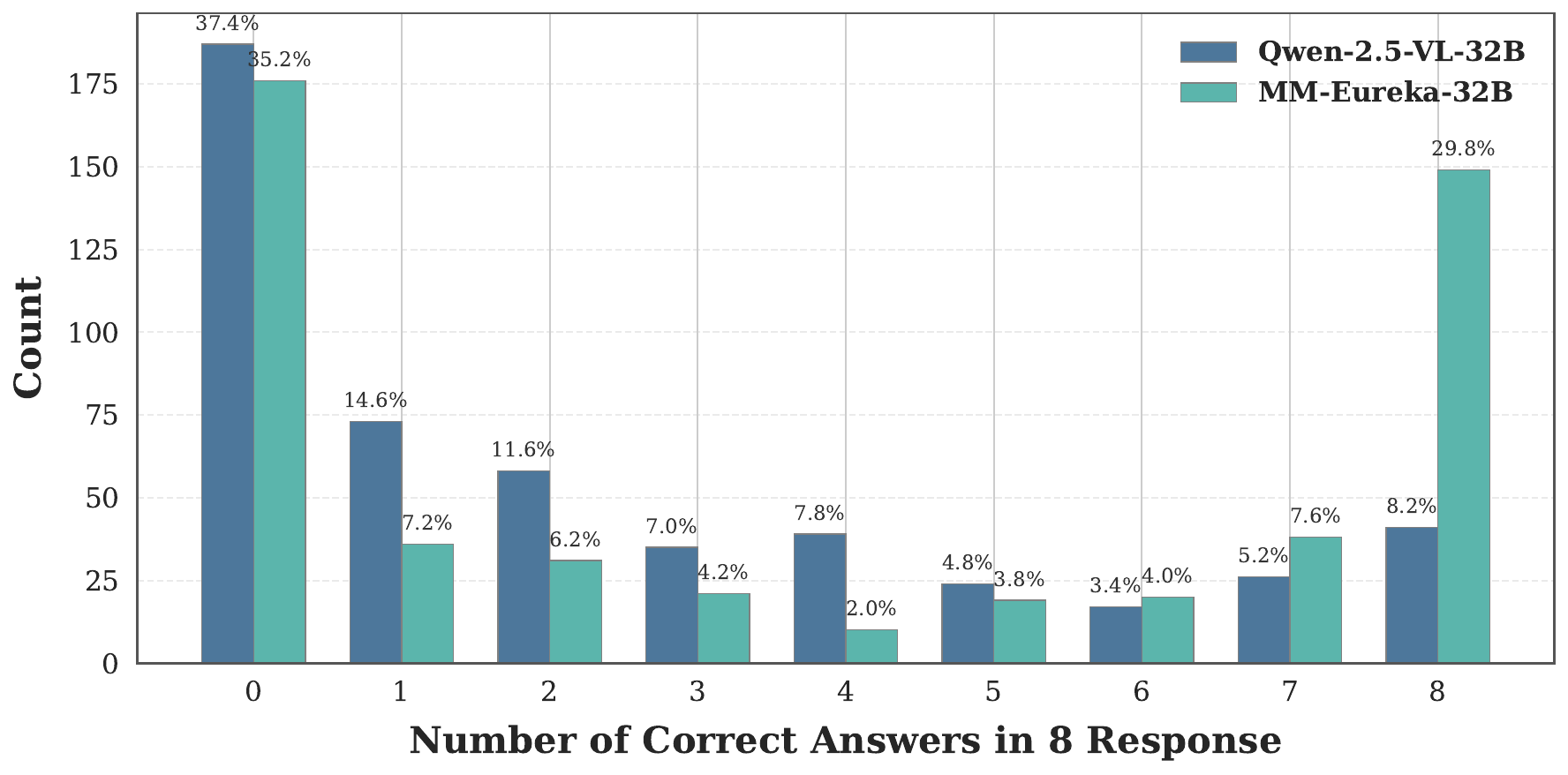}
    \caption{Distribution of correct answers across 8 responses from Qwen-2.5-VL-32B and MM-Eureka-32B on \textit{Mathematics}. }
    \label{fig: comparison of math testset}
\end{figure}

Figure~\ref{fig: comparison of math testset} presents the distribution of correct answers across 8 responses from the Qwen-2.5-VL-32B and MM-Eureka-32B models on the {Mathematics} dataset. It is evident that, for problems that were initially answered correctly at least once, Eureka shows a significant improvement in accuracy. However, the number of problems that initially had zero correct answers remains nearly unchanged. 
We hypothesize that this improvement is due to Eureka’s improved reasoning capabilities, developed through RL training, allowing it to better leverage existing knowledge and thus improve the accuracy on partially understood problems. Nonetheless, relying solely on reasoning is insufficient to solve problems the model originally cannot answer, which is why the accuracy for problems with zero correct answers remains almost unchanged.

Figure~\ref{fig:qualitative mathematics and physics} shows the responses from Qwen-2.5-VL-32B and MM-Eureka-32B to the same question from {Mathematics} (the full version is provided in Appendix). From the responses, we observe that although Qwen possesses the necessary knowledge to answer the question, it still struggles to provide the correct answer. In contrast, Eureka is able to better reason with the available knowledge and derive the correct answer. 
Table~\ref{tab:training_methods_comparison} further illustrates that, despite Eureka's training set being focused only on mathematics, it also demonstrates a notable improvement in answering questions from other domains such as physics, chemistry, and biology.

Our experimental results to some extent support the view that knowledge and reasoning in LLMs/VLMs can be decoupled, and to some extent indicate that learning knowledge and learning reasoning can be conducted separately during the training process.
Therefore, a key direction for future research is to explore how to generalize the reasoning capabilities, developed from data like mathematics—where logic is clear, responses are structured, and answers are verifiable—into broader, more general domains.

\subsection{RL generalizes better than SFT}

We maintain consistent settings with our RL training to compare different post-training strategies including SFT and COT SFT. Using the ms-swift framework \citep{zhao2024swiftascalablelightweightinfrastructure}, we conduct both SFT and COT SFT training for 10 epochs with identical data. As shown in Table \ref{tab:training_methods_comparison}, results demonstrate that RL exhibits superior generalization compared to SFT or COT SFT approaches, particularly with more significant improvements on OOD test sets such as Physics, Chemistry, and Biology. Furthermore, RL enhances the model's reasoning capabilities more effectively—while SFT and COT SFT fail to substantially improve performance on mathematics and physics problems, RL training increases the model's scores in mathematics and physics by 12.8 and 10.8 points respectively.

\begin{table}[h]
    \centering
    \caption{Performance comparison of different training methods on MMK12. In terms of both enhancing mathematical capabilities and generalizing to other disciplines, RL significantly outperforms SFT or COT SFT.}\vspace{5pt}
    \scalebox{1.0}{
    \begin{tabular}{lccccc}
        \toprule
        \textbf{Model} & \textbf{Mathematics} & \textbf{Physics} & \textbf{Chemistry} & \textbf{Biology} & \textbf{Avg.} \\
        \midrule
        Qwen-2.5-VL-7B & 58.4 & 45.4 & 56.4 & 54.0 & 53.5 \\
        + SFT & 56.6 & 50.0 & 63.2 & 61.2 & 57.7 \\
        + COT SFT & 59.2 & 46.0 & 62.2 & 61.2 & 57.1 \\
        + RL & \textbf{71.2} & \textbf{56.2} &\textbf{65.2}& \textbf{65.0} & \textbf{64.5} \\
        \bottomrule
    \end{tabular}
    }
    \label{tab:training_methods_comparison}
\end{table}


\subsection{Two-Stage Training} \label{sec:twostageabli}

\begin{figure*}[t]
    \centering
    \includegraphics[width=1.0\linewidth]{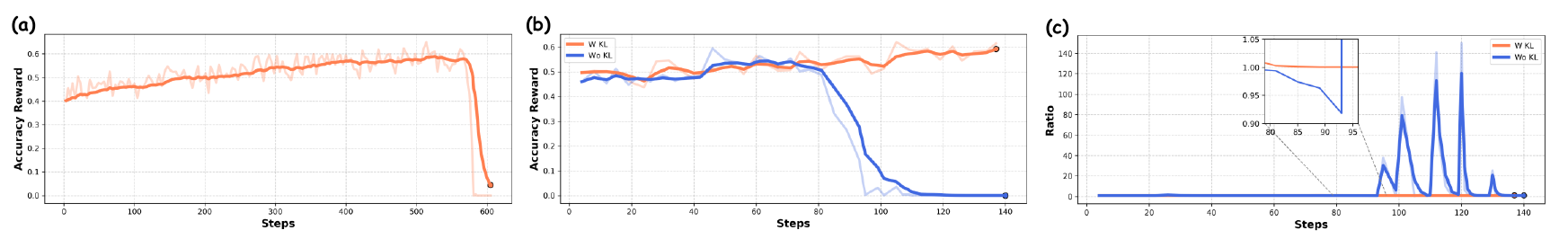}
    \vspace{-0.1in}
    \caption{(a) MM-Eureka-32B tends to experience sudden training collapse during RL training, manifested as accuracy reward approaching zero. (b) During the second training phase, adding KL divergence leads to more stable training with steadily increasing accuracy rewards. (c) The stabilizing effect of KL divergence in the second phase may be attributed to the stabilization of the policy ratio, as ratio instability often leads to training collapse.}
    \label{fig:twostage-log}
\end{figure*}

In Section \ref{sec:twostage}, we introduce a two-stage RL training strategy. Stage 1 employs K12 data to enhance general knowledge reasoning capabilities, while stage 2 utilizes Geo3k data to address specific performance gaps. Additionally, we omit the KL divergence in stage one to accelerate training and then incorporate it in stage two to maintain training stability.

As shown in Figure \ref{fig:twostage-log} (a), continuous training without KL divergence often leads to sudden model collapse. We identify this phenomenon results from ratio instability (excessive updates in the new policy). As shown in Figures \ref{fig:twostage-log} (b) and (c), which present training metrics with and without the use of KL divergence in the second stage, a sharp fluctuation in the ratio is observed when the model begins to collapse. To mitigate this issue of ratio instability, we introduce a KL divergence term during the second stage of training, which helps to stabilize the learning dynamics. 

Furthermore, Table \ref{tab:twostage_math_benchmarks} demonstrates that after the first training stage, despite overall performance improvements, capabilities in certain specific domains actually decline, as evidenced by MathVista's geometry problem solving(GPS) score dropping from $74.0$ to $56.7$. To address this, we implement second-stage training using Geo3k, which yields further improvements across multiple mathematical benchmarks.

\begin{table}[h]
    \centering
    \caption{Performance comparison across multimodal mathematical benchmarks of the two stage training strategy. The second phase of training enhances various capabilities of the model following the first phase, with particularly significant improvements in geometric reasoning abilities. } \vspace{5pt}
    \scalebox{0.85}{
    \begin{tabular}{lcccccc}
        \toprule
        \textbf{Model} & \textbf{Data Scale} & \textbf{MathVista} & \textbf{MathVista(GPS)} & \textbf{MathVerse} & \textbf{Mathvision} & \textbf{WeMath}  \\
        \midrule
        Stage1 & 15k  & 72.0 & 56.7 & 56.4 & 33.6 & \textbf{73.5} \\
        Stage2 & 2k  & \textbf{74.8} &  \textbf{70.2}  & \textbf{56.5} & \textbf{34.4} & 73.4  \\
        \bottomrule
    \end{tabular}
    }
    \label{tab:twostage_math_benchmarks}
\end{table}

\section{Conclusion}
Our work aims at developing a powerful model for multimodal reasoning. To achieve this, we first propose MMK12, which is a high quality multimodal reasoning mathematical dataset collected by ourselves. Moreover, we utilize online filter and propose a two-stage training strategy to make the rule-based reinforcement learning training more stable and efficient. Finally, we propose MM-Eureka-7B and MM-Eureka-32B, both are the top performers within the similar size models in multimodal reasoning tasks. Especially for multidisciplinary reasoning tasks, MM-Eureka-32B surpasses almost all open-source models and closed-source models, which is only slightly behind o1.

\section{Acknowledgements}
We acknowledge the outstanding open-source contributions from vLLM, LMM-R1, and OpenRLHF. We also extend our gratitude to DeepSeek-R1, InternVL, and QwenVL for their open-source techniques and base models, which have enabled us to further our exploration.


\bibliography{iclr2025_conference}

\begin{thebibliography}{42}
\providecommand{\natexlab}[1]{#1}
\providecommand{\url}[1]{\texttt{#1}}
\expandafter\ifx\csname urlstyle\endcsname\relax
  \providecommand{\doi}[1]{doi: #1}\else
  \providecommand{\doi}{doi: \begingroup \urlstyle{rm}\Url}\fi

\bibitem[Ahmadian et~al.(2024)Ahmadian, Cremer, Gall{\'{e}}, Fadaee, Kreutzer, Pietquin, {\"{U}}st{\"{u}}n, and Hooker]{RLOOAhmadianCGFKPUH24}
Arash Ahmadian, Chris Cremer, Matthias Gall{\'{e}}, Marzieh Fadaee, Julia Kreutzer, Olivier Pietquin, Ahmet {\"{U}}st{\"{u}}n, and Sara Hooker.
\newblock Back to basics: Revisiting reinforce-style optimization for learning from human feedback in llms.
\newblock In \emph{Proceedings of the 62nd Annual Meeting of the Association for Computational Linguistics (Volume 1: Long Papers), {ACL} 2024}, pp.\  12248--12267. Association for Computational Linguistics, 2024.

\bibitem[{Anthropic}(2024)]{claude3.7sonnet}
{Anthropic}.
\newblock Claude 3.7 sonnet.
\newblock \url{https://www.anthropic.com/claude}, 2024.
\newblock Accessed: 2025-04-14.

\bibitem[Bai et~al.(2023)Bai, Bai, Yang, Wang, Tan, Wang, Lin, Zhou, and Zhou]{bai2023qwenvlversatilevisionlanguagemodel}
Jinze Bai, Shuai Bai, Shusheng Yang, Shijie Wang, Sinan Tan, Peng Wang, Junyang Lin, Chang Zhou, and Jingren Zhou.
\newblock Qwen-vl: A versatile vision-language model for understanding, localization, text reading, and beyond, 2023.
\newblock URL \url{https://arxiv.org/abs/2308.12966}.

\bibitem[Chen et~al.(2025)Chen, Li, Zhao, Song, and Vinci]{chen2025r1v}
Liang Chen, Lei Li, Haozhe Zhao, Yifan Song, and Vinci.
\newblock R1-v: Reinforcing super generalization ability in vision-language models with less than \$3.
\newblock \url{https://github.com/Deep-Agent/R1-V}, 2025.
\newblock Accessed: 2025-02-02.

\bibitem[Chen et~al.(2024)Chen, Wang, Cao, Liu, Gao, Cui, Zhu, Ye, Tian, Liu, et~al.]{chen2024expanding}
Zhe Chen, Weiyun Wang, Yue Cao, Yangzhou Liu, Zhangwei Gao, Erfei Cui, Jinguo Zhu, Shenglong Ye, Hao Tian, Zhaoyang Liu, et~al.
\newblock Expanding performance boundaries of open-source multimodal models with model, data, and test-time scaling.
\newblock \emph{arXiv preprint arXiv:2412.05271}, 2024.

\bibitem[Cui et~al.(2025)Cui, Yuan, Wang, Wang, Li, He, Fan, Yu, Xu, Chen, Yuan, Chen, Zhang, Lv, Wang, Yao, Han, Peng, Cheng, Liu, Sun, Zhou, and Ding]{cui2025processreinforcementimplicitrewards}
Ganqu Cui, Lifan Yuan, Zefan Wang, Hanbin Wang, Wendi Li, Bingxiang He, Yuchen Fan, Tianyu Yu, Qixin Xu, Weize Chen, Jiarui Yuan, Huayu Chen, Kaiyan Zhang, Xingtai Lv, Shuo Wang, Yuan Yao, Xu~Han, Hao Peng, Yu~Cheng, Zhiyuan Liu, Maosong Sun, Bowen Zhou, and Ning Ding.
\newblock Process reinforcement through implicit rewards, 2025.
\newblock URL \url{https://arxiv.org/abs/2502.01456}.

\bibitem[DeepSeek-AI et~al.(2025)DeepSeek-AI, Guo, Yang, Zhang, Song, Zhang, Xu, Zhu, Ma, Wang, Bi, Zhang, Yu, Wu, Wu, Gou, Shao, Li, Gao, Liu, Xue, Wang, Wu, Feng, Lu, Zhao, Deng, Zhang, Ruan, Dai, Chen, Ji, Li, Lin, Dai, Luo, Hao, Chen, Li, Zhang, Bao, Xu, Wang, Ding, Xin, Gao, Qu, Li, Guo, Li, Wang, Chen, Yuan, Qiu, Li, Cai, Ni, Liang, Chen, Dong, Hu, Gao, Guan, Huang, Yu, Wang, Zhang, Zhao, Wang, Zhang, Xu, Xia, Zhang, Zhang, Tang, Li, Wang, Li, Tian, Huang, Zhang, Wang, Chen, Du, Ge, Zhang, Pan, Wang, Chen, Jin, Chen, Lu, Zhou, Chen, Ye, Wang, Yu, Zhou, Pan, Li, Zhou, Wu, Ye, Yun, Pei, Sun, Wang, Zeng, Zhao, Liu, Liang, Gao, Yu, Zhang, Xiao, An, Liu, Wang, Chen, Nie, Cheng, Liu, Xie, Liu, Yang, Li, Su, Lin, Li, Jin, Shen, Chen, Sun, Wang, Song, Zhou, Wang, Shan, Li, Wang, Wei, Zhang, Xu, Li, Zhao, Sun, Wang, Yu, Zhang, Shi, Xiong, He, Piao, Wang, Tan, Ma, Liu, Guo, Ou, Wang, Gong, Zou, He, Xiong, Luo, You, Liu, Zhou, Zhu, Xu, Huang, Li, Zheng, Zhu, Ma, Tang, Zha, Yan, Ren, Ren, Sha, Fu, Xu, Xie, Zhang,
  Hao, Ma, Yan, Wu, Gu, Zhu, Liu, Li, Xie, Song, Pan, Huang, Xu, Zhang, and Zhang]{deepseekai2025}
DeepSeek-AI, Daya Guo, Dejian Yang, Haowei Zhang, Junxiao Song, Ruoyu Zhang, Runxin Xu, Qihao Zhu, Shirong Ma, Peiyi Wang, Xiao Bi, Xiaokang Zhang, Xingkai Yu, Yu~Wu, Z.~F. Wu, Zhibin Gou, Zhihong Shao, Zhuoshu Li, Ziyi Gao, Aixin Liu, Bing Xue, Bingxuan Wang, Bochao Wu, Bei Feng, Chengda Lu, Chenggang Zhao, Chengqi Deng, Chenyu Zhang, Chong Ruan, Damai Dai, Deli Chen, Dongjie Ji, Erhang Li, Fangyun Lin, Fucong Dai, Fuli Luo, Guangbo Hao, Guanting Chen, Guowei Li, H.~Zhang, Han Bao, Hanwei Xu, Haocheng Wang, Honghui Ding, Huajian Xin, Huazuo Gao, Hui Qu, Hui Li, Jianzhong Guo, Jiashi Li, Jiawei Wang, Jingchang Chen, Jingyang Yuan, Junjie Qiu, Junlong Li, J.~L. Cai, Jiaqi Ni, Jian Liang, Jin Chen, Kai Dong, Kai Hu, Kaige Gao, Kang Guan, Kexin Huang, Kuai Yu, Lean Wang, Lecong Zhang, Liang Zhao, Litong Wang, Liyue Zhang, Lei Xu, Leyi Xia, Mingchuan Zhang, Minghua Zhang, Minghui Tang, Meng Li, Miaojun Wang, Mingming Li, Ning Tian, Panpan Huang, Peng Zhang, Qiancheng Wang, Qinyu Chen, Qiushi Du, Ruiqi Ge, Ruisong
  Zhang, Ruizhe Pan, Runji Wang, R.~J. Chen, R.~L. Jin, Ruyi Chen, Shanghao Lu, Shangyan Zhou, Shanhuang Chen, Shengfeng Ye, Shiyu Wang, Shuiping Yu, Shunfeng Zhou, Shuting Pan, S.~S. Li, Shuang Zhou, Shaoqing Wu, Shengfeng Ye, Tao Yun, Tian Pei, Tianyu Sun, T.~Wang, Wangding Zeng, Wanjia Zhao, Wen Liu, Wenfeng Liang, Wenjun Gao, Wenqin Yu, Wentao Zhang, W.~L. Xiao, Wei An, Xiaodong Liu, Xiaohan Wang, Xiaokang Chen, Xiaotao Nie, Xin Cheng, Xin Liu, Xin Xie, Xingchao Liu, Xinyu Yang, Xinyuan Li, Xuecheng Su, Xuheng Lin, X.~Q. Li, Xiangyue Jin, Xiaojin Shen, Xiaosha Chen, Xiaowen Sun, Xiaoxiang Wang, Xinnan Song, Xinyi Zhou, Xianzu Wang, Xinxia Shan, Y.~K. Li, Y.~Q. Wang, Y.~X. Wei, Yang Zhang, Yanhong Xu, Yao Li, Yao Zhao, Yaofeng Sun, Yaohui Wang, Yi~Yu, Yichao Zhang, Yifan Shi, Yiliang Xiong, Ying He, Yishi Piao, Yisong Wang, Yixuan Tan, Yiyang Ma, Yiyuan Liu, Yongqiang Guo, Yuan Ou, Yuduan Wang, Yue Gong, Yuheng Zou, Yujia He, Yunfan Xiong, Yuxiang Luo, Yuxiang You, Yuxuan Liu, Yuyang Zhou, Y.~X. Zhu,
  Yanhong Xu, Yanping Huang, Yaohui Li, Yi~Zheng, Yuchen Zhu, Yunxian Ma, Ying Tang, Yukun Zha, Yuting Yan, Z.~Z. Ren, Zehui Ren, Zhangli Sha, Zhe Fu, Zhean Xu, Zhenda Xie, Zhengyan Zhang, Zhewen Hao, Zhicheng Ma, Zhigang Yan, Zhiyu Wu, Zihui Gu, Zijia Zhu, Zijun Liu, Zilin Li, Ziwei Xie, Ziyang Song, Zizheng Pan, Zhen Huang, Zhipeng Xu, Zhongyu Zhang, and Zhen Zhang.
\newblock Deepseek-r1: Incentivizing reasoning capability in llms via reinforcement learning, 2025.
\newblock URL \url{https://arxiv.org/abs/2501.12948}.

\bibitem[Deng et~al.(2024)Deng, Liu, Li, Luo, Wu, Zhang, Lyu, Zhang, Zhang, Ding, et~al.]{deng2024r}
Linger Deng, Yuliang Liu, Bohan Li, Dongliang Luo, Liang Wu, Chengquan Zhang, Pengyuan Lyu, Ziyang Zhang, Gang Zhang, Errui Ding, et~al.
\newblock R-cot: Reverse chain-of-thought problem generation for geometric reasoning in large multimodal models.
\newblock \emph{arXiv preprint arXiv:2410.17885}, 2024.

\bibitem[Deng et~al.(2025)Deng, Bansal, Yin, Peng, Wang, and Chang]{deng2025openvlthinker}
Yihe Deng, Hritik Bansal, Fan Yin, Nanyun Peng, Wei Wang, and Kai-Wei Chang.
\newblock Openvlthinker: An early exploration to complex vision-language reasoning via iterative self-improvement, 2025.
\newblock URL \url{https://arxiv.org/abs/2503.17352}.

\bibitem[Duan et~al.(2024)Duan, Yang, Qiao, Fang, Chen, Liu, Dong, Zang, Zhang, Wang, et~al.]{duan2024vlmevalkit}
Haodong Duan, Junming Yang, Yuxuan Qiao, Xinyu Fang, Lin Chen, Yuan Liu, Xiaoyi Dong, Yuhang Zang, Pan Zhang, Jiaqi Wang, et~al.
\newblock Vlmevalkit: An open-source toolkit for evaluating large multi-modality models.
\newblock In \emph{Proceedings of the 32nd ACM international conference on multimedia}, pp.\  11198--11201, 2024.

\bibitem[Gao et~al.(2022)Gao, Schulman, and Hilton]{gao2022scalinglawsrewardmodel}
Leo Gao, John Schulman, and Jacob Hilton.
\newblock Scaling laws for reward model overoptimization, 2022.
\newblock URL \url{https://arxiv.org/abs/2210.10760}.

\bibitem[Gui \& Ren(2025)Gui and Ren]{gui2025adora}
Lujun Gui and Qingnan Ren.
\newblock Training reasoning model with dynamic advantage estimation on reinforcement learning.
\newblock \url{https://www.notion.so/Training_Reasoning_Model_with_Dynamic_Advantage_Estimation_on_Reinforcement_Learning_1a830cc0904681fa9df3e076b6557a3e}, 2025.
\newblock Notion Blog.

\bibitem[Guo et~al.(2024)Guo, Zheng, Bai, Li, Wang, Zhu, Li, Neubig, Chen, and Yue]{guo2024mammothvlelicitingmultimodalreasoning}
Jarvis Guo, Tuney Zheng, Yuelin Bai, Bo~Li, Yubo Wang, King Zhu, Yizhi Li, Graham Neubig, Wenhu Chen, and Xiang Yue.
\newblock Mammoth-vl: Eliciting multimodal reasoning with instruction tuning at scale, 2024.
\newblock URL \url{https://arxiv.org/abs/2412.05237}.

\bibitem[He et~al.(2024)He, Luo, Bai, Hu, Thai, Shen, Hu, Han, Huang, Zhang, Liu, Qi, Liu, and Sun]{he2024olympiadbenchchallengingbenchmarkpromoting}
Chaoqun He, Renjie Luo, Yuzhuo Bai, Shengding Hu, Zhen~Leng Thai, Junhao Shen, Jinyi Hu, Xu~Han, Yujie Huang, Yuxiang Zhang, Jie Liu, Lei Qi, Zhiyuan Liu, and Maosong Sun.
\newblock Olympiadbench: A challenging benchmark for promoting agi with olympiad-level bilingual multimodal scientific problems, 2024.
\newblock URL \url{https://arxiv.org/abs/2402.14008}.

\bibitem[Hu(2025)]{hu2025reinforce++}
Jian Hu.
\newblock Reinforce++: A simple and efficient approach for aligning large language models.
\newblock \emph{arXiv preprint arXiv:2501.03262}, 2025.

\bibitem[Hu et~al.(2024)Hu, Wu, Zhu, Xianyu, Wang, Zhang, and Cao]{hu2024openrlhf}
Jian Hu, Xibin Wu, Zilin Zhu, Xianyu, Weixun Wang, Dehao Zhang, and Yu~Cao.
\newblock Openrlhf: An easy-to-use, scalable and high-performance rlhf framework, 2024.
\newblock URL \url{https://arxiv.org/abs/2405.11143}.

\bibitem[Hurst et~al.(2024)Hurst, Lerer, Goucher, Perelman, Ramesh, Clark, Ostrow, Welihinda, Hayes, Radford, et~al.]{hurst2024gpt}
Aaron Hurst, Adam Lerer, Adam~P Goucher, Adam Perelman, Aditya Ramesh, Aidan Clark, AJ~Ostrow, Akila Welihinda, Alan Hayes, Alec Radford, et~al.
\newblock Gpt-4o system card.
\newblock \emph{arXiv preprint arXiv:2410.21276}, 2024.

\bibitem[Kool et~al.(2019)Kool, van Hoof, and Welling]{RLOOKoolHW19a}
Wouter Kool, Herke van Hoof, and Max Welling.
\newblock Buy 4 {REINFORCE} samples, get a baseline for free!
\newblock In \emph{Deep Reinforcement Learning Meets Structured Prediction, {ICLR} 2019 Workshop}. OpenReview.net, 2019.

\bibitem[Kwon et~al.(2023)Kwon, Li, Zhuang, Sheng, Zheng, Yu, Gonzalez, Zhang, and Stoica]{kwon2023efficient}
Woosuk Kwon, Zhuohan Li, Siyuan Zhuang, Ying Sheng, Lianmin Zheng, Cody~Hao Yu, Joseph~E. Gonzalez, Hao Zhang, and Ion Stoica.
\newblock Efficient memory management for large language model serving with pagedattention.
\newblock In \emph{Proceedings of the ACM SIGOPS 29th Symposium on Operating Systems Principles}, 2023.

\bibitem[Lu et~al.(2021)Lu, Gong, Jiang, Qiu, Huang, Liang, and Zhu]{lu2021inter}
Pan Lu, Ran Gong, Shibiao Jiang, Liang Qiu, Siyuan Huang, Xiaodan Liang, and Song-Chun Zhu.
\newblock Inter-gps: Interpretable geometry problem solving with formal language and symbolic reasoning.
\newblock \emph{arXiv preprint arXiv:2105.04165}, 2021.

\bibitem[Lu et~al.(2024)Lu, Bansal, Xia, Liu, Li, Hajishirzi, Cheng, Chang, Galley, and Gao]{lu2024mathvistaevaluatingmathematicalreasoning}
Pan Lu, Hritik Bansal, Tony Xia, Jiacheng Liu, Chunyuan Li, Hannaneh Hajishirzi, Hao Cheng, Kai-Wei Chang, Michel Galley, and Jianfeng Gao.
\newblock Mathvista: Evaluating mathematical reasoning of foundation models in visual contexts, 2024.
\newblock URL \url{https://arxiv.org/abs/2310.02255}.

\bibitem[OpenAI(2024)]{openai2024o1}
OpenAI.
\newblock Introducing openai o1.
\newblock \url{https://openai.com/o1/}, 2024.
\newblock Accessed: 2024-10-02.

\bibitem[Ouyang et~al.(2022)Ouyang, Wu, Jiang, Almeida, Wainwright, Mishkin, Zhang, Agarwal, Slama, Ray, Schulman, Hilton, Kelton, Miller, Simens, Askell, Welinder, Christiano, Leike, and Lowe]{ouyang2022traininglanguagemodelsfollow}
Long Ouyang, Jeff Wu, Xu~Jiang, Diogo Almeida, Carroll~L. Wainwright, Pamela Mishkin, Chong Zhang, Sandhini Agarwal, Katarina Slama, Alex Ray, John Schulman, Jacob Hilton, Fraser Kelton, Luke Miller, Maddie Simens, Amanda Askell, Peter Welinder, Paul Christiano, Jan Leike, and Ryan Lowe.
\newblock Training language models to follow instructions with human feedback, 2022.
\newblock URL \url{https://arxiv.org/abs/2203.02155}.

\bibitem[Peng et~al.(2024)Peng, Fu, Gao, Zhong, Fu, and Tang]{peng2024multimath}
Shuai Peng, Di~Fu, Liangcai Gao, Xiuqin Zhong, Hongguang Fu, and Zhi Tang.
\newblock Multimath: Bridging visual and mathematical reasoning for large language models.
\newblock \emph{arXiv preprint arXiv:2409.00147}, 2024.

\bibitem[Peng et~al.(2025)Peng, Zhang, Geng, and Yang]{peng2025lmmr1}
YingZhe Peng, Gongrui Zhang, Xin Geng, and Xu~Yang.
\newblock Lmm-r1.
\newblock \url{https://github.com/TideDra/lmm-r1}, 2025.
\newblock Accessed: 2025-02-13.

\bibitem[Qiao et~al.(2024)Qiao, Tan, Dong, Wu, Sun, Song, Gongque, Lei, Wei, Zhang, Qiao, Zhang, Zong, Xu, Diao, Bao, Li, and Zhang]{qiao2024we}
Runqi Qiao, Qiuna Tan, Guanting Dong, Minhui Wu, Chong Sun, Xiaoshuai Song, Zhuoma Gongque, Shanglin Lei, Zhe Wei, Miaoxuan Zhang, Runfeng Qiao, Yifan Zhang, Xiao Zong, Yida Xu, Muxi Diao, Zhimin Bao, Chen Li, and Honggang Zhang.
\newblock We-math: Does your large multimodal model achieve human-like mathematical reasoning?
\newblock \emph{CoRR}, abs/2407.01284, 2024.
\newblock \doi{10.48550/ARXIV.2407.01284}.
\newblock URL \url{https://doi.org/10.48550/arXiv.2407.01284}.

\bibitem[Radford et~al.(2019)Radford, Wu, Child, Luan, Amodei, Sutskever, et~al.]{radford2019language}
Alec Radford, Jeffrey Wu, Rewon Child, David Luan, Dario Amodei, Ilya Sutskever, et~al.
\newblock Language models are unsupervised multitask learners.
\newblock \emph{OpenAI blog}, 1\penalty0 (8):\penalty0 9, 2019.

\bibitem[Rafailov et~al.(2023)Rafailov, Sharma, Mitchell, Manning, Ermon, and Finn]{rafailov2023direct}
Rafael Rafailov, Archit Sharma, Eric Mitchell, Christopher~D Manning, Stefano Ermon, and Chelsea Finn.
\newblock Direct preference optimization: Your language model is secretly a reward model.
\newblock \emph{Advances in Neural Information Processing Systems}, 36:\penalty0 53728--53741, 2023.

\bibitem[Schulman et~al.(2017{\natexlab{a}})Schulman, Wolski, Dhariwal, Radford, and Klimov]{PPO}
John Schulman, Filip Wolski, Prafulla Dhariwal, Alec Radford, and Oleg Klimov.
\newblock Proximal policy optimization algorithms, 2017{\natexlab{a}}.
\newblock URL \url{https://arxiv.org/abs/1707.06347}.

\bibitem[Schulman et~al.(2017{\natexlab{b}})Schulman, Wolski, Dhariwal, Radford, and Klimov]{schulman2017proximal}
John Schulman, Filip Wolski, Prafulla Dhariwal, Alec Radford, and Oleg Klimov.
\newblock Proximal policy optimization algorithms.
\newblock \emph{arXiv preprint arXiv:1707.06347}, 2017{\natexlab{b}}.

\bibitem[Sutton et~al.(1998)Sutton, Barto, et~al.]{sutton1998reinforcement}
Richard~S Sutton, Andrew~G Barto, et~al.
\newblock \emph{Reinforcement learning: An introduction}, volume~1.
\newblock MIT press Cambridge, 1998.

\bibitem[Tan et~al.(2025)Tan, Ji, Hao, Lin, Wang, Wang, and Zhang]{tan2025reasonrftreinforcementfinetuningvisual}
Huajie Tan, Yuheng Ji, Xiaoshuai Hao, Minglan Lin, Pengwei Wang, Zhongyuan Wang, and Shanghang Zhang.
\newblock Reason-rft: Reinforcement fine-tuning for visual reasoning, 2025.
\newblock URL \url{https://arxiv.org/abs/2503.20752}.

\bibitem[Team et~al.(2023)Team, Anil, Borgeaud, Alayrac, Yu, Soricut, Schalkwyk, Dai, Hauth, Millican, et~al.]{team2023gemini}
Gemini Team, Rohan Anil, Sebastian Borgeaud, Jean-Baptiste Alayrac, Jiahui Yu, Radu Soricut, Johan Schalkwyk, Andrew~M Dai, Anja Hauth, Katie Millican, et~al.
\newblock Gemini: a family of highly capable multimodal models.
\newblock \emph{arXiv preprint arXiv:2312.11805}, 2023.

\bibitem[Team et~al.(2025{\natexlab{a}})Team, Du, Gao, Xing, Jiang, Chen, Li, Xiao, Du, Liao, Tang, Wang, Zhang, Yuan, Lu, Tang, Sung, Wei, Lai, Guo, Zhu, Ding, Hu, Yang, Zhang, Yao, Zhao, Lu, Li, Yu, Gao, Zheng, Yuan, Chen, Guo, Su, Wang, Zhao, Zhang, Liu, Yan, Wu, Shi, Ye, Yu, Dong, Zhang, Ma, Pan, Gong, Liu, Ma, Wei, Cao, Huang, Jiang, Gao, Xiong, He, Huang, Wu, He, Wei, Jia, Wu, Xu, Zu, Zhou, Pan, Charles, Li, Hu, Liu, Chen, Wang, Liu, Qin, Liu, Yang, Bao, Du, Wu, Wang, Zhou, Wang, Li, Zhu, Zhang, Wang, Yang, Huang, Huang, Xu, and Yang]{kimiteam2025kimik15scalingreinforcement}
Kimi Team, Angang Du, Bofei Gao, Bowei Xing, Changjiu Jiang, Cheng Chen, Cheng Li, Chenjun Xiao, Chenzhuang Du, Chonghua Liao, Chuning Tang, Congcong Wang, Dehao Zhang, Enming Yuan, Enzhe Lu, Fengxiang Tang, Flood Sung, Guangda Wei, Guokun Lai, Haiqing Guo, Han Zhu, Hao Ding, Hao Hu, Hao Yang, Hao Zhang, Haotian Yao, Haotian Zhao, Haoyu Lu, Haoze Li, Haozhen Yu, Hongcheng Gao, Huabin Zheng, Huan Yuan, Jia Chen, Jianhang Guo, Jianlin Su, Jianzhou Wang, Jie Zhao, Jin Zhang, Jingyuan Liu, Junjie Yan, Junyan Wu, Lidong Shi, Ling Ye, Longhui Yu, Mengnan Dong, Neo Zhang, Ningchen Ma, Qiwei Pan, Qucheng Gong, Shaowei Liu, Shengling Ma, Shupeng Wei, Sihan Cao, Siying Huang, Tao Jiang, Weihao Gao, Weimin Xiong, Weiran He, Weixiao Huang, Wenhao Wu, Wenyang He, Xianghui Wei, Xianqing Jia, Xingzhe Wu, Xinran Xu, Xinxing Zu, Xinyu Zhou, Xuehai Pan, Y.~Charles, Yang Li, Yangyang Hu, Yangyang Liu, Yanru Chen, Yejie Wang, Yibo Liu, Yidao Qin, Yifeng Liu, Ying Yang, Yiping Bao, Yulun Du, Yuxin Wu, Yuzhi Wang, Zaida Zhou,
  Zhaoji Wang, Zhaowei Li, Zhen Zhu, Zheng Zhang, Zhexu Wang, Zhilin Yang, Zhiqi Huang, Zihao Huang, Ziyao Xu, and Zonghan Yang.
\newblock Kimi k1.5: Scaling reinforcement learning with llms, 2025{\natexlab{a}}.
\newblock URL \url{https://arxiv.org/abs/2501.12599}.

\bibitem[Team et~al.(2025{\natexlab{b}})Team, Du, Yin, Xing, Qu, Wang, Chen, Zhang, Du, Wei, Wang, Zhang, Du, Wang, Yuan, Lu, Li, Sung, Wei, Lai, Zhu, Ding, Hu, Yang, Zhang, Wu, Yao, Lu, Wang, Gao, Zheng, Li, Su, Wang, Deng, Qiu, Xie, Wang, Liu, Yan, Ouyang, Chen, Sui, Yu, Dong, Dong, Xu, Cheng, Gu, Zhou, Liu, Cao, Yu, Song, Bai, Song, He, Huang, Xu, Yuan, Yao, Wu, Zu, Zhou, Wang, Charles, Zhong, Li, Hu, Chen, Wang, Liu, Miao, Qin, Chen, Bao, Wang, Kang, Liu, Du, Wu, Wang, Yan, Zhou, Li, Jiang, Zhang, Yang, Huang, Huang, Zhao, and Chen]{kimiteam2025kimivltechnicalreport}
Kimi Team, Angang Du, Bohong Yin, Bowei Xing, Bowen Qu, Bowen Wang, Cheng Chen, Chenlin Zhang, Chenzhuang Du, Chu Wei, Congcong Wang, Dehao Zhang, Dikang Du, Dongliang Wang, Enming Yuan, Enzhe Lu, Fang Li, Flood Sung, Guangda Wei, Guokun Lai, Han Zhu, Hao Ding, Hao Hu, Hao Yang, Hao Zhang, Haoning Wu, Haotian Yao, Haoyu Lu, Heng Wang, Hongcheng Gao, Huabin Zheng, Jiaming Li, Jianlin Su, Jianzhou Wang, Jiaqi Deng, Jiezhong Qiu, Jin Xie, Jinhong Wang, Jingyuan Liu, Junjie Yan, Kun Ouyang, Liang Chen, Lin Sui, Longhui Yu, Mengfan Dong, Mengnan Dong, Nuo Xu, Pengyu Cheng, Qizheng Gu, Runjie Zhou, Shaowei Liu, Sihan Cao, Tao Yu, Tianhui Song, Tongtong Bai, Wei Song, Weiran He, Weixiao Huang, Weixin Xu, Xiaokun Yuan, Xingcheng Yao, Xingzhe Wu, Xinxing Zu, Xinyu Zhou, Xinyuan Wang, Y.~Charles, Yan Zhong, Yang Li, Yangyang Hu, Yanru Chen, Yejie Wang, Yibo Liu, Yibo Miao, Yidao Qin, Yimin Chen, Yiping Bao, Yiqin Wang, Yongsheng Kang, Yuanxin Liu, Yulun Du, Yuxin Wu, Yuzhi Wang, Yuzi Yan, Zaida Zhou, Zhaowei Li, Zhejun
  Jiang, Zheng Zhang, Zhilin Yang, Zhiqi Huang, Zihao Huang, Zijia Zhao, and Ziwei Chen.
\newblock {Kimi-VL} technical report, 2025{\natexlab{b}}.
\newblock URL \url{https://arxiv.org/abs/2504.07491}.

\bibitem[Team(2024)]{qvq-72b-preview}
Qwen Team.
\newblock Qvq: To see the world with wisdom, December 2024.
\newblock URL \url{https://qwenlm.github.io/blog/qvq-72b-preview/}.

\bibitem[Wang et~al.(2024{\natexlab{a}})Wang, Pan, Shi, Lu, Zhan, and Li]{wang2024measuringmultimodalmathematicalreasoning}
Ke~Wang, Junting Pan, Weikang Shi, Zimu Lu, Mingjie Zhan, and Hongsheng Li.
\newblock Measuring multimodal mathematical reasoning with math-vision dataset, 2024{\natexlab{a}}.
\newblock URL \url{https://arxiv.org/abs/2402.14804}.

\bibitem[Wang et~al.(2024{\natexlab{b}})Wang, Chen, Wang, Cao, Liu, Gao, Zhu, Zhu, Lu, Qiao, and Dai]{wang2024enhancingreasoningabilitymultimodal}
Weiyun Wang, Zhe Chen, Wenhai Wang, Yue Cao, Yangzhou Liu, Zhangwei Gao, Jinguo Zhu, Xizhou Zhu, Lewei Lu, Yu~Qiao, and Jifeng Dai.
\newblock Enhancing the reasoning ability of multimodal large language models via mixed preference optimization, 2024{\natexlab{b}}.
\newblock URL \url{https://arxiv.org/abs/2411.10442}.

\bibitem[Yang et~al.(2025)Yang, He, Pan, Jiang, Deng, Yang, Lu, Yin, Rao, Zhu, Zhang, and Chen]{yang2025r1onevision}
Yi~Yang, Xiaoxuan He, Hongkun Pan, Xiyan Jiang, Yan Deng, Xingtao Yang, Haoyu Lu, Dacheng Yin, Fengyun Rao, Minfeng Zhu, Bo~Zhang, and Wei Chen.
\newblock R1-onevision: Advancing generalized multimodal reasoning through cross-modal formalization.
\newblock \emph{arXiv preprint arXiv:2503.10615}, 2025.

\bibitem[Zhang et~al.(2024{\natexlab{a}})Zhang, Jiang, Zhang, Lin, Guo, Qiu, Zhou, Lu, Chang, Gao, and Li]{zhang2024mathversedoesmultimodalllm}
Renrui Zhang, Dongzhi Jiang, Yichi Zhang, Haokun Lin, Ziyu Guo, Pengshuo Qiu, Aojun Zhou, Pan Lu, Kai-Wei Chang, Peng Gao, and Hongsheng Li.
\newblock Mathverse: Does your multi-modal llm truly see the diagrams in visual math problems?, 2024{\natexlab{a}}.
\newblock URL \url{https://arxiv.org/abs/2403.14624}.

\bibitem[Zhang et~al.(2024{\natexlab{b}})Zhang, Wei, Jiang, Guo, Li, Zhang, Tong, Liu, Zhou, Wei, et~al.]{zhang2024mavis}
Renrui Zhang, Xinyu Wei, Dongzhi Jiang, Ziyu Guo, Shicheng Li, Yichi Zhang, Chengzhuo Tong, Jiaming Liu, Aojun Zhou, Bin Wei, et~al.
\newblock Mavis: Mathematical visual instruction tuning with an automatic data engine.
\newblock \emph{arXiv preprint arXiv:2407.08739}, 2024{\natexlab{b}}.

\bibitem[Zhao et~al.(2024)Zhao, Huang, Hu, Wang, Mao, Zhang, Jiang, Wu, Ai, Wang, Zhou, and Chen]{zhao2024swiftascalablelightweightinfrastructure}
Yuze Zhao, Jintao Huang, Jinghan Hu, Xingjun Wang, Yunlin Mao, Daoze Zhang, Zeyinzi Jiang, Zhikai Wu, Baole Ai, Ang Wang, Wenmeng Zhou, and Yingda Chen.
\newblock Swift:a scalable lightweight infrastructure for fine-tuning, 2024.
\newblock URL \url{https://arxiv.org/abs/2408.05517}.

\end{thebibliography}
\bibliographystyle{iclr2025_conference}

\newpage

\appendix



\section{Appendix: Qualitative Analysis}\label{appendix:qualitative}

\begin{figure}[ht!]
    \centering
    {\fontsize{8pt}{9pt}\selectfont \begin{tcolorbox}[
        enhanced,
        sharp corners,
        colback=white,
        frame hidden,
        boxrule=0pt,
        skin=enhancedmiddle,
        overlay unbroken={
                \draw[decorate, decoration={random steps, segment length=20pt, amplitude=1pt}, line width=0.75pt]
                (frame.south west) -- (frame.north west) --
                (frame.north east) -- (frame.south east) -- cycle;
            },
        overlay first={
                \draw[decorate, decoration={random steps, segment length=20pt, amplitude=1pt}, line width=0.75pt]
                (frame.south west) -- (frame.north west) --
                (frame.north east) -- (frame.south east) -- cycle;
            },
        overlay middle={
                \draw[decorate, decoration={random steps, segment length=20pt, amplitude=1pt}, line width=0.75pt]
                (frame.south west) -- (frame.north west) --
                (frame.north east) -- (frame.south east) -- cycle;
            },
        overlay last={
                \draw[decorate, decoration={random steps, segment length=20pt, amplitude=1pt}, line width=0.75pt]
                (frame.south west) -- (frame.north west) --
                (frame.north east) -- (frame.south east) -- cycle;
            },
    ]
    \begin{tabular}{@{} m{0.82\linewidth} m{0.14\linewidth} @{}}
        \textbf{Question:} \newline
        As shown in the figure, in the straight square column $ABCD - A_{1}B_{1}C_{1}D_{1}$, $\angle ADC = 90^\circ$, and $AA_1 = AD = DC = 1$, $M \in$ plane $ABCD$, when $D_1M \perp$ plane $A_1C_1D$, $D_1M = $\_\_\_.\newline \newline
        \textbf{Answer:} $\sqrt{3}$
         &
        \centering
        \includegraphics[width=\linewidth]{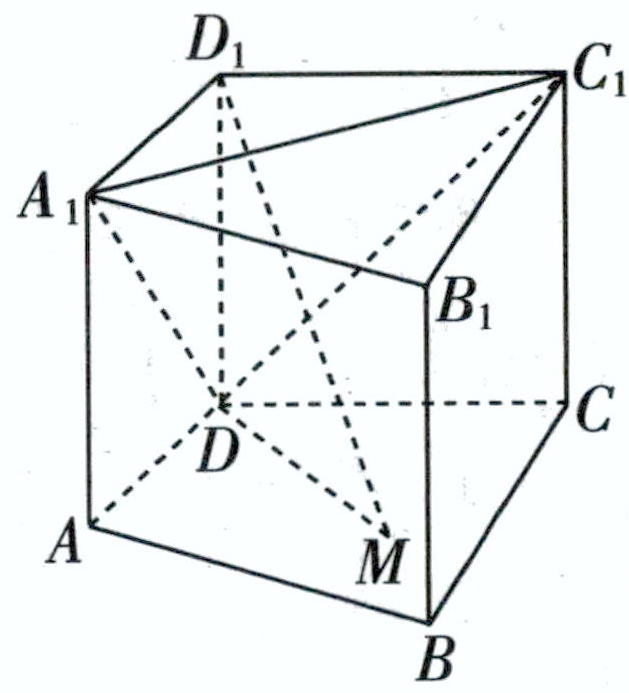}
    \end{tabular}
\end{tcolorbox}

\begin{tcolorbox}[
        enhanced,
        sharp corners,
        colback=white,
        frame hidden,
        boxrule=0pt,
        skin=enhancedmiddle,
        before upper={
                \raisebox{-0.4em}{\includegraphics[width=1.5em]{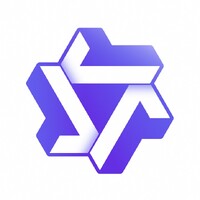}}~\textbf{Qwen2.5-VL-32B-Instruct}\\[0.5em]
            },
        overlay unbroken={
                \draw[decorate, decoration={random steps, segment length=20pt, amplitude=1pt}, line width=0.75pt]
                (frame.south west) -- (frame.north west) --
                (frame.north east) -- (frame.south east) -- cycle;
            },
        overlay first={
                \draw[decorate, decoration={random steps, segment length=20pt, amplitude=1pt}, line width=0.75pt]
                (frame.south west) -- (frame.north west) --
                (frame.north east) -- (frame.south east) -- cycle;
            },
        overlay middle={
                \draw[decorate, decoration={random steps, segment length=20pt, amplitude=1pt}, line width=0.75pt]
                (frame.south west) -- (frame.north west) --
                (frame.north east) -- (frame.south east) -- cycle;
            },
        overlay last={
                \draw[decorate, decoration={random steps, segment length=20pt, amplitude=1pt}, line width=0.75pt]
                (frame.south west) -- (frame.north west) --
                (frame.north east) -- (frame.south east) -- cycle;
            },
    ]
    To solve the problem, we need to determine the length of \( D_1M \), where \( M \) is a point in the plane \( ABCD \) such that \( D_1M \perp \) plane \( A_1C_1D \).

    \dots

    \textbf{2. Coordinates of Points:} Place the prism in a 3D Cartesian coordinate system:
    \[
        A = (0, 0, 0), \quad B = (1, 0, 0), \quad C = (1, 1, 0), \quad D = (0, 1, 0)
    \]
    \[
        A_1 = (0, 0, 1), \quad B_1 = (1, 0, 1), \quad C_1 = (1, 1, 1), \quad D_1 = (0, 1, 1)
    \]
    Let \( M = (x, y, 0) \), where \( M \) is in the plane \( ABCD \).

    \textbf{3. Vector Representation:} $ \overrightarrow{D_1M} = (x - 0, y - 1, 0 - 1) = (x, y - 1, -1)$

    \textbf{4. Condition for Perpendicularity:} \textcolor{red!75!black!}{For \( D_1M \perp \) plane \( A_1C_1D \), the vector \( \overrightarrow{D_1M} \) must be perpendicular to the normal vector of the plane \( A_1C_1D \).}

    First, find two non-parallel vectors in the plane \( A_1C_1D \):

    \dots
    
    Therefore, the length of \( D_1M \) is \( \boxed{ \frac{ \sqrt{6} }{2 } } \).

\end{tcolorbox}

\begin{tcolorbox}[
        enhanced,
        sharp corners,
        colback=white,
        frame hidden,
        boxrule=0pt,
        skin=enhancedmiddle,
        before upper={
                \raisebox{-0.4em}{\includegraphics[width=1.5em]{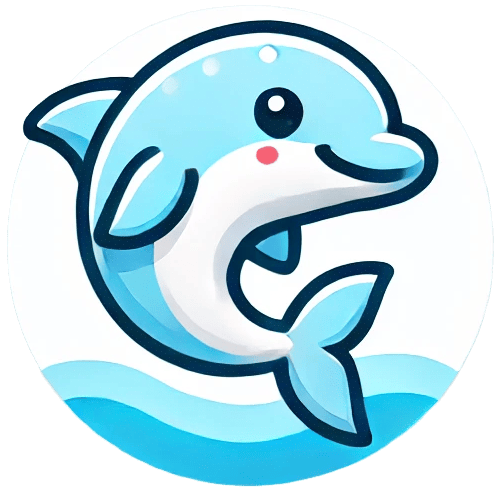}}~\textbf{MM-EUREKA-32B}\\[0.5em]
            },
        overlay unbroken={
                \draw[decorate, decoration={random steps, segment length=20pt, amplitude=1pt}, line width=0.75pt]
                (frame.south west) -- (frame.north west) --
                (frame.north east) -- (frame.south east) -- cycle;
            },
        overlay first={
                \draw[decorate, decoration={random steps, segment length=20pt, amplitude=1pt}, line width=0.75pt]
                (frame.south west) -- (frame.north west) --
                (frame.north east) -- (frame.south east) -- cycle;
            },
        overlay middle={
                \draw[decorate, decoration={random steps, segment length=20pt, amplitude=1pt}, line width=0.75pt]
                (frame.south west) -- (frame.north west) --
                (frame.north east) -- (frame.south east) -- cycle;
            },
        overlay last={
                \draw[decorate, decoration={random steps, segment length=20pt, amplitude=1pt}, line width=0.75pt]
                (frame.south west) -- (frame.north west) --
                (frame.north east) -- (frame.south east) -- cycle;
            },
    ]
    To solve the problem, we need to determine the length of $D_1M$ when $D_1M \perp$ plane $A_1C_1D$. Let's break the problem into steps:

    \dots

    \textbf{2. Determination of the plane $A_1C_1D$:}
    \begin{itemize}[leftmargin=2em, itemsep=0.5pt, topsep=0.5pt]
        \item The points $A_1 = (1, 0, 1)$, $C_1 = (0, 1, 1)$, and $D = (0, 0, 0)$ lie in the plane $A_1C_1D$.
        \item To find the normal vector to the plane $A_1C_1D$, we use the cross product of two vectors in the plane:
              \begin{align*}
                  \overrightarrow{A_1D} = (1 - 0, 0 - 0, 1 - 0) = (1, 0, 1), \quad
                  \overrightarrow{C_1D} = (0 - 0, 1 - 0, 1 - 0) = (0, 1, 1).
              \end{align*}
        \item The normal vector $\mathbf{n}$ to the plane is given by the cross product:
              \[
                  \mathbf{n} = \overrightarrow{A_1D} \times \overrightarrow{C_1D} =
                  \begin{vmatrix}
                      \mathbf{i} & \mathbf{j} & \mathbf{k} \\
                      1          & 0          & 1          \\
                      0          & 1          & 1
                  \end{vmatrix}
                  = \mathbf{i}(0 \cdot 1 - 1 \cdot 1)
                  - \mathbf{j}(1 \cdot 1 - 1 \cdot 0)
                  + \mathbf{k}(1 \cdot 1 - 0 \cdot 0)
                  = (-1, -1, 1).
              \]
        \item Thus, the normal vector to the plane $A_1C_1D$ is $\mathbf{n} = (-1, -1, 1)$.
    \end{itemize}

    \textbf{3. Condition for $D_1M \perp$ plane $A_1C_1D$:}
    \begin{itemize}[leftmargin=2em, itemsep=0.5pt, topsep=0.5pt]
        \item \textcolor{green!45!black!}{Since $D_1M \perp$ plane $A_1C_1D$, the vector $\overrightarrow{D_1M}$ must be parallel to the normal vector $\mathbf{n} = (-1, -1, 1)$.}
        \item Let $M = (x, y, 0)$ be a point in the plane $ABCD$ (since $M \in$ plane $ABCD$, the $z$-coordinate of $M$ is 0).
        \item The vector $\overrightarrow{D_1M} = (x - 0, y - 0, 0 - 1) = (x, y, -1)$.
    \end{itemize}

    \dots

    \textbf{4. Calculation of $D_1M$:}
    \begin{itemize}[leftmargin=2em, itemsep=0.5pt, topsep=0.5pt]
        \item The vector $\overrightarrow{D_1M} = (1 - 0, 1 - 0, 0 - 1) = (1, 1, -1)$.
        \item The length of $D_1M$ is given by the magnitude of $\overrightarrow{D_1M}$: $ |\overrightarrow{D_1M}| = \sqrt{1^2 + 1^2 + (-1)^2} = \sqrt{1 + 1 + 1} = \sqrt{3}. $
    \end{itemize}

    \textless answer\textgreater The answer is $ \boxed{\sqrt{3}} $. \textless /answer\textgreater
\end{tcolorbox}}
    \caption{Example from mathematics. \textcolor{red!75!black!}{Red} highlights Qwen2.5-VL-32B-Instruct's reasoning errors. \textcolor{green!45!black!}{Green} shows MM-Eureka-32B's correct reasoning.}
    \label{fig:qualitative_math}
\end{figure}

\newpage

\begin{figure}[ht!]
    \centering
    {\fontsize{8pt}{9pt}\selectfont \begin{tcolorbox}[
        enhanced,
        breakable,
        sharp corners,
        colback=white,
        frame hidden,
        boxrule=0pt,
        skin=enhancedmiddle,
        overlay unbroken={
                \draw[decorate, decoration={random steps, segment length=20pt, amplitude=1pt}, line width=0.75pt]
                (frame.south west) -- (frame.north west) --
                (frame.north east) -- (frame.south east) -- cycle;
            },
        overlay first={
                \draw[decorate, decoration={random steps, segment length=20pt, amplitude=1pt}, line width=0.75pt]
                (frame.south west) -- (frame.north west) --
                (frame.north east) -- (frame.south east) -- cycle;
            },
        overlay middle={
                \draw[decorate, decoration={random steps, segment length=20pt, amplitude=1pt}, line width=0.75pt]
                (frame.south west) -- (frame.north west) --
                (frame.north east) -- (frame.south east) -- cycle;
            },
        overlay last={
                \draw[decorate, decoration={random steps, segment length=20pt, amplitude=1pt}, line width=0.75pt]
                (frame.south west) -- (frame.north west) --
                (frame.north east) -- (frame.south east) -- cycle;
            },
    ]
    \begin{tabular}{@{} m{0.82\linewidth} m{0.14\linewidth} @{}}
        \textbf{Question:} \newline
        As shown in the figure, a lightweight spring is placed vertically on a horizontal surface with a mass of $6kg$ object $A$ resting on its upper end, in a stationary state. Now, a mass of $4kg$ object $B$ is gently placed on top of object $A$, and the spring remains within its elastic limit. Taking $g$ as $10m/s^{2}$, at the moment $B$ is placed, the magnitude of the force between $A$ and $B$ is $(\quad)$\vspace{0.5em}
        A. $8N$ ~ ~
        B. $18N$ ~ ~
        C. $24N$ ~ ~
        D. $40N$\newline \newline
        \textbf{Answer:} $C$
         &
        \centering
        \includegraphics[width=\linewidth]{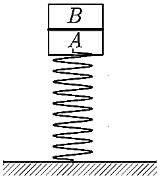}
    \end{tabular}
\end{tcolorbox}

\begin{tcolorbox}[
        enhanced,
        breakable,
        sharp corners,
        colback=white,
        frame hidden,
        boxrule=0pt,
        skin=enhancedmiddle,
        before upper={
                \raisebox{-0.4em}{\includegraphics[width=1.5em]{qwen-logo.jpeg}}~\textbf{Qwen2.5-VL-32B-Instruct}\\[0.5em]
            },
        overlay unbroken={
                \draw[decorate, decoration={random steps, segment length=20pt, amplitude=1pt}, line width=0.75pt]
                (frame.south west) -- (frame.north west) --
                (frame.north east) -- (frame.south east) -- cycle;
            },
        overlay first={
                \draw[decorate, decoration={random steps, segment length=20pt, amplitude=1pt}, line width=0.75pt]
                (frame.south west) -- (frame.north west) --
                (frame.north east) -- (frame.south east) -- cycle;
            },
        overlay middle={
                \draw[decorate, decoration={random steps, segment length=20pt, amplitude=1pt}, line width=0.75pt]
                (frame.south west) -- (frame.north west) --
                (frame.north east) -- (frame.south east) -- cycle;
            },
        overlay last={
                \draw[decorate, decoration={random steps, segment length=20pt, amplitude=1pt}, line width=0.75pt]
                (frame.south west) -- (frame.north west) --
                (frame.north east) -- (frame.south east) -- cycle;
            },
    ]
    To solve this problem, we need to determine the force between objects $A$ and $B$ at the moment $B$ is placed on top of $A$. Let's break it down step by step.

    \dots

    \textbf{4. Force Analysis at the Moment $B$ is Placed:}

    \begin{itemize}[leftmargin=2em, itemsep=0.5pt, topsep=0.5pt]
        \item At the moment $B$ is placed, the spring has not yet had time to extend further to balance the new total weight. The spring force remains $60\, \text{N}$ (the initial force that balances the weight of $A$).
        \item \textcolor{red!75!black!}{The net force on $A$ is the sum of the spring force and the force from $B$:}{\color{red!75!black!}$ F_s + F_B = m_A g. $}
    
        \item Here, $F_s = 60\, \text{N}$ and $m_A g = 60\, \text{N}$, so: $60\, \text{N} + F_B = 60\,\text{N}. $
        \item Solving for $F_B$, we get: $ F_B = 0\, \text{N}. $
    \end{itemize}

    \textbf{5. Re-evaluate the Situation:}

    \begin{itemize}[leftmargin=2em, itemsep=0.5pt, topsep=0.5pt]
        \item The above analysis is not correct because we need to consider the instantaneous force between $A$ and $B$ due to the sudden addition of $B$.
        \item \textcolor{red!75!black!}{The force between $A$ and $B$ is the force $B$ exerts on $A$, which is the weight of $B$: $ F_B = m_B g = 4\, \text{kg} \cdot 10\, \text{m/s}^2 = 40\, \text{N}. $}
    \end{itemize}

      \[
          \boxed{40}
      \]

\end{tcolorbox}

\begin{tcolorbox}[
        enhanced,
        breakable,
        sharp corners,
        colback=white,
        frame hidden,
        boxrule=0pt,
        skin=enhancedmiddle,
        before upper={
                \raisebox{-0.4em}{\includegraphics[width=1.5em]{mm-eureka-logo.png}}~\textbf{MM-EUREKA-32B}\\[0.5em]
            },
        overlay unbroken={
                \draw[decorate, decoration={random steps, segment length=20pt, amplitude=1pt}, line width=0.75pt]
                (frame.south west) -- (frame.north west) --
                (frame.north east) -- (frame.south east) -- cycle;
            },
        overlay first={
                \draw[decorate, decoration={random steps, segment length=20pt, amplitude=1pt}, line width=0.75pt]
                (frame.south west) -- (frame.north west) --
                (frame.north east) -- (frame.south east) -- cycle;
            },
        overlay middle={
                \draw[decorate, decoration={random steps, segment length=20pt, amplitude=1pt}, line width=0.75pt]
                (frame.south west) -- (frame.north west) --
                (frame.north east) -- (frame.south east) -- cycle;
            },
        overlay last={
                \draw[decorate, decoration={random steps, segment length=20pt, amplitude=1pt}, line width=0.75pt]
                (frame.south west) -- (frame.north west) --
                (frame.north east) -- (frame.south east) -- cycle;
            },
    ]
    To solve this problem, we need to determine the force between objects $A$ and $B$ at the moment $B$ is placed on top of $A$. Let's break the problem into steps:

    \dots

    \textbf{4. Applying Newton's Second Law to $A$}

    \begin{align*}
        F_{\text{net}} & = F_B + F_A - F_{\text{spring}} \\
                       & = F_B + 60 - 60 = F_B.
    \end{align*}

    The net force on $A$ causes an upward acceleration $a$, which is the same as the acceleration of $B$ because they are in contact.

    \textbf{5. Analyzing the Forces on $B$}

    \begin{itemize}[leftmargin=2em, itemsep=0.5pt, topsep=0.5pt]
        \item The forces on $B$ are:
              \begin{itemize}[leftmargin=2em, itemsep=0.5pt, topsep=0.5pt]
                  \item The weight of $B$: $F_B = 40 \, \text{N}$ (downward),
                  \item The force from $A$: $F_{\text{contact}}$ (upward).
              \end{itemize}
    \end{itemize}

    \textbf{6. Applying Newton's Second Law to $B$}

    \[
        F_B - F_{\text{contact}} = m_B a.
    \]

    \textbf{7. Finding the Acceleration $a$}

    \begin{itemize}[leftmargin=2em, itemsep=0.5pt, topsep=0.5pt]
        \item The total mass of the system is $ m_{\text{total}} = m_A + m_B = 6 + 4 = 10 \, \text{kg}. $
        \item The net force on the system is the extra weight from $B$: $ F_{\text{net, total}} = 40 \, \text{N}. $
        \item Using Newton's Second Law for the entire system: $ F_{\text{net, total}} = m_{\text{total}} a \Rightarrow 40 = 10a \Rightarrow a = 4 \, \text{m/s}^2. $
    \end{itemize}

    \textbf{8. Finding the Contact Force $F_{\text{contact}}$}

    \textcolor{green!45!black!}{\begin{align*}
            F_B - F_{\text{contact}} & = m_B a                     \\
            40 - F_{\text{contact}}  & = 4 \cdot 4 = 16            \\
            F_{\text{contact}}       & = 40 - 16 = 24 \, \text{N}.
        \end{align*}}

    \textless answer\textgreater The answer is $ \boxed{C} $. \textless /answer\textgreater
\end{tcolorbox}}
    \caption{Example from physics. \textcolor{red!75!black!}{Red} highlights Qwen2.5-VL-32B-Instruct's reasoning errors. \textcolor{green!45!black!}{Green} shows MM-Eureka-32B's correct reasoning.}
    \label{fig:qualitative_phy}
\end{figure}

\newpage

\begin{figure}[ht!]
    \centering
    {\fontsize{8pt}{9pt}\selectfont \begin{tcolorbox}[
        enhanced,
        breakable,
        sharp corners,
        colback=white,
        frame hidden,
        boxrule=0pt,
        skin=enhancedmiddle,
        overlay unbroken={
                \draw[decorate, decoration={random steps, segment length=20pt, amplitude=1pt}, line width=0.75pt]
                (frame.south west) -- (frame.north west) --
                (frame.north east) -- (frame.south east) -- cycle;
            },
        overlay first={
                \draw[decorate, decoration={random steps, segment length=20pt, amplitude=1pt}, line width=0.75pt]
                (frame.south west) -- (frame.north west) --
                (frame.north east) -- (frame.south east) -- cycle;
            },
        overlay middle={
                \draw[decorate, decoration={random steps, segment length=20pt, amplitude=1pt}, line width=0.75pt]
                (frame.south west) -- (frame.north west) --
                (frame.north east) -- (frame.south east) -- cycle;
            },
        overlay last={
                \draw[decorate, decoration={random steps, segment length=20pt, amplitude=1pt}, line width=0.75pt]
                (frame.south west) -- (frame.north west) --
                (frame.north east) -- (frame.south east) -- cycle;
            },
    ]
    \begin{tabular}{@{} m{0.62\linewidth} m{0.34\linewidth} @{}}
        \textbf{Question:} \newline
        At room temperature, a $\rm 0.100\ mol/L\ \rm NaOH$ solution is used to titrate $\rm 25.00 \ mL \ 0.100\ mol/L\ \rm H_2C_2O_4$ solution, and the resulting titration curve is shown in the figure (the volume of the mixed solution can be considered as the sum of the volumes of the solutions before mixing). Which of the following statements is incorrect? $(\quad)$\newline
        \textbf{A}. In the solution at point X: $c{\rm (Na^+)}>c{\rm (HC_2O_4^-)}>c{\rm (C_2O_4^{2-})}$\newline
        \textbf{B}. Observing the solution at point Y indicates that the degree of ionization of $\rm HC_2O_4^-$ is greater than the degree of hydrolysis of $\rm C_2O_4^{2-}$\newline
        \textbf{C}. In the solution at point Z: $c{\rm (Na^+)}>c{\rm (C_2O_4^{2-})}>c{\rm (HC_2O_4^-)}>c{\rm (OH^-)}>c{\rm (H^+)} $\newline
        \textbf{D}. Throughout the titration process: $c{\rm (Na^+)}+c{\rm (H_2C_2O_4)}+c{\rm (HC_2O_4^-)}+c{\rm (C_2O_4^{2-})}=\rm 0.100 mol/L$ \newline \newline
        \textbf{Answer:} $C$
         &
        \centering
        \includegraphics[width=\linewidth]{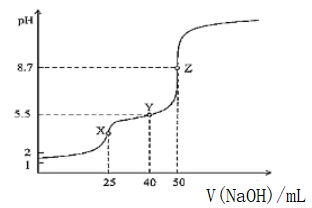}
    \end{tabular}
\end{tcolorbox}

\begin{tcolorbox}[
        enhanced,
        breakable,
        sharp corners,
        colback=white,
        frame hidden,
        boxrule=0pt,
        skin=enhancedmiddle,
        before upper={
                \raisebox{-0.4em}{\includegraphics[width=1.5em]{qwen-logo.jpeg}}~\textbf{Qwen2.5-VL-32B-Instruct}\\[0.5em]
            },
        overlay unbroken={
                \draw[decorate, decoration={random steps, segment length=20pt, amplitude=1pt}, line width=0.75pt]
                (frame.south west) -- (frame.north west) --
                (frame.north east) -- (frame.south east) -- cycle;
            },
        overlay first={
                \draw[decorate, decoration={random steps, segment length=20pt, amplitude=1pt}, line width=0.75pt]
                (frame.south west) -- (frame.north west) --
                (frame.north east) -- (frame.south east) -- cycle;
            },
        overlay middle={
                \draw[decorate, decoration={random steps, segment length=20pt, amplitude=1pt}, line width=0.75pt]
                (frame.south west) -- (frame.north west) --
                (frame.north east) -- (frame.south east) -- cycle;
            },
        overlay last={
                \draw[decorate, decoration={random steps, segment length=20pt, amplitude=1pt}, line width=0.75pt]
                (frame.south west) -- (frame.north west) --
                (frame.north east) -- (frame.south east) -- cycle;
            },
    ]
    \textbf{1. Understanding the Reaction:}
    \begin{itemize}[leftmargin=2em, itemsep=0.5pt, topsep=0.5pt]
        \item The titration is between a strong base (0.100 M NaO) and a weak dicarboxylic acid (0.100 M H$_2$CO$_4$).
        \item The reaction steps are:
              \begin{enumerate}
                  \item H$_2$CO$_4$ + NaO $\rightarrow$ NaHC$_2$O$_4$ + H$_2$O
                  \item NaHC$_2$O$_4$ + NaO $\rightarrow$ Na$_2$CO$_4$ + H$_2$O
              \end{enumerate}
    \end{itemize}

    \textbf{2. Key Points on the pH Curve:}
    \begin{itemize}[leftmargin=2em, itemsep=0.5pt, topsep=0.5pt]
        \item \textcolor{red!75!black!}{\textbf{Point X (pH $\approx$ 2):} This is the initial point where only H$_2$CO$_4$ is present. The solution is acidic.}
        \item \textcolor{red!75!black!}{\textbf{Point Y (pH $\approx$ 5.5):} This is the first equivalence point where all H$_2$CO$_4$ has been neutralized to NaHC$_2$O$_4$ (solutions of NaHC$_2$O$_4$ are acidic due to the protonation of the conjugate base).}
        \item \textbf{Point Z (pH $\approx$ 8.7):} This is the second equivalence point where all NaHC$_2$O$_4$ has been neutralized to Na$_2$CO$_4$ (solutions of Na$_2$CO$_4$ are basic due to the deprotonation of the conjugate base).
    \end{itemize}

    \dots

    \textbf{4. Final Answer:} The only option that is \textbf{incorrect} is Option \boxed{A}.
\end{tcolorbox}

\begin{tcolorbox}[
        enhanced,
        breakable,
        sharp corners,
        colback=white,
        frame hidden,
        boxrule=0pt,
        skin=enhancedmiddle,
        before upper={
                \raisebox{-0.4em}{\includegraphics[width=1.5em]{mm-eureka-logo.png}}~\textbf{MM-EUREKA-32B}\\[0.5em]
            },
        overlay unbroken={
                \draw[decorate, decoration={random steps, segment length=20pt, amplitude=1pt}, line width=0.75pt]
                (frame.south west) -- (frame.north west) --
                (frame.north east) -- (frame.south east) -- cycle;
            },
        overlay first={
                \draw[decorate, decoration={random steps, segment length=20pt, amplitude=1pt}, line width=0.75pt]
                (frame.south west) -- (frame.north west) --
                (frame.north east) -- (frame.south east) -- cycle;
            },
        overlay middle={
                \draw[decorate, decoration={random steps, segment length=20pt, amplitude=1pt}, line width=0.75pt]
                (frame.south west) -- (frame.north west) --
                (frame.north east) -- (frame.south east) -- cycle;
            },
        overlay last={
                \draw[decorate, decoration={random steps, segment length=20pt, amplitude=1pt}, line width=0.75pt]
                (frame.south west) -- (frame.north west) --
                (frame.north east) -- (frame.south east) -- cycle;
            },
    ]
    To solve this problem, we need to analyze the titration of a weak dicarboxylic acid $ \mathrm{H_2C_2O_4} $ (oxalate) with a strong base $ \mathrm{NaOH} $. The titration process involves multiple steps, and we must consider the equilibrium and ionic concentrations at different points on the titration curve. Let's break the problem into parts and solve it step by step.

    \textbf{1. Understanding the titration process:}

    $ \mathrm{H_2C_2O_4} $ is a weak dicarboxylic acid, which undergoes two steps of dissociation:
    \begin{align*}
        \mathrm{H_2C_2O_4} & \rightleftharpoons \mathrm{H^+ + HC_2O_4^-}   \\
        \mathrm{HC_2O_4^-} & \rightleftharpoons \mathrm{H^+ + C_2O_4^{2-}}
    \end{align*}

    The titration is carried out with $ \mathrm{NaOH} $, a strong base, which will neutralize the $ \mathrm{H^+} $ ions step by step.

    \textbf{2. Key points on the titration curve:}

    \textcolor{green!45!black!}{\textbf{Point X:} 25 mL of $ \mathrm{NaOH} $ added.}
    \begin{itemize}[leftmargin=2em, itemsep=0.5pt, topsep=0.5pt]
        \item \textcolor{green!45!black!}{At this point, the first equivalence point is reached. All $ \mathrm{H_2C_2O_4} $ is converted to $ \mathrm{NaHC_2O_4} $ (sodium hydrogen oxalate).}
        \item \textcolor{green!45!black!}{The solution contains $ \mathrm{NaHC_2O_4} $, and the pH is around 2, indicating that $ \mathrm{HC_2O_4^-} $ is a weak acid.}
    \end{itemize}

    \textcolor{green!45!black!}{\textbf{Point Y:} 40 mL of $ \mathrm{NaOH} $ added.}
    \begin{itemize}[leftmargin=2em, itemsep=0.5pt, topsep=0.5pt]
        \item \textcolor{green!45!black!}{This is between the first and second equivalence points. The solution contains a mixture of $ \mathrm{NaHC_2O_4} $ and $ \mathrm{Na_2C_2O_4} $.}
        \item \textcolor{green!45!black!}{The pH is 5.5, indicating that the solution is slightly acidic due to the dissociation of $ \mathrm{HC_2O_4^-} $ being greater than the hydrolysis of $ \mathrm{C_2O_4^{2-}} $.}
    \end{itemize}

    \textbf{Point Z:} 50 mL of $ \mathrm{NaOH} $ added.
    \begin{itemize}[leftmargin=2em, itemsep=0.5pt, topsep=0.5pt]
        \item At this point, the second equivalence point is reached. All $ \mathrm{H_2C_2O_4} $ is converted to $ \mathrm{Na_2C_2O_4} $ (sodium oxalate).
        \item The solution contains $ \mathrm{Na_2C_2O_4} $, and the pH is 8.7, indicating that $ \mathrm{C_2O_4^{2-}} $ is a strong base.
    \end{itemize}

    \dots

    \textless answer\textgreater The answer is $ \boxed{C} $. \textless /answer\textgreater
\end{tcolorbox}}
    \caption{Example from chemistry. \textcolor{red!75!black!}{Red} highlights Qwen2.5-VL-32B-Instruct's reasoning errors. \textcolor{green!45!black!}{Green} shows MM-Eureka-32B's correct reasoning.}
    \label{fig:qualitative_chem}
\end{figure}

\newpage

\begin{figure}[ht!]
    \centering
    {\fontsize{8pt}{9pt}\selectfont \begin{tcolorbox}[
        enhanced,
        sharp corners,
        colback=white,
        frame hidden,
        boxrule=0pt,
        skin=enhancedmiddle,
        overlay unbroken={
                \draw[decorate, decoration={random steps, segment length=20pt, amplitude=1pt}, line width=0.75pt]
                (frame.south west) -- (frame.north west) --
                (frame.north east) -- (frame.south east) -- cycle;
            },
        overlay first={
                \draw[decorate, decoration={random steps, segment length=20pt, amplitude=1pt}, line width=0.75pt]
                (frame.south west) -- (frame.north west) --
                (frame.north east) -- (frame.south east) -- cycle;
            },
        overlay middle={
                \draw[decorate, decoration={random steps, segment length=20pt, amplitude=1pt}, line width=0.75pt]
                (frame.south west) -- (frame.north west) --
                (frame.north east) -- (frame.south east) -- cycle;
            },
        overlay last={
                \draw[decorate, decoration={random steps, segment length=20pt, amplitude=1pt}, line width=0.75pt]
                (frame.south west) -- (frame.north west) --
                (frame.north east) -- (frame.south east) -- cycle;
            },
    ]
    \begin{tabular}{@{} m{0.77\linewidth} m{0.19\linewidth} @{}}
        \textbf{Question:} \newline
        Scientists isolated gene A from a certain organism and dissociated it into two single strands. One of these strands was hybridized with the messenger RNA of gene A, resulting in the pattern shown in the image. An unreasonable explanation for this is $(\quad)$\newline
        \textbf{A}. Positions 1, 3, 5, and 7 are connected through complementary base pairing\newline
        \textbf{B}. Heating to a certain temperature can separate the two strands\newline
        \textbf{C}. The diagram contains 8 types of nucleotides, 5 types of bases, and 2 types of base pairs\newline
        \textbf{D}. Since fragments 2, 2, and 6 cannot bind with the messenger RNA, fragments 2, 4, and 6 belong to non-gene segments of the DNA molecule\newline\newline
        \textbf{Answer:} $D$
         &
        \centering
        \includegraphics[width=\linewidth]{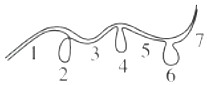}
    \end{tabular}
\end{tcolorbox}

\begin{tcolorbox}[
        enhanced,
        sharp corners,
        colback=white,
        frame hidden,
        boxrule=0pt,
        skin=enhancedmiddle,
        before upper={
                \raisebox{-0.4em}{\includegraphics[width=1.5em]{qwen-logo.jpeg}}~\textbf{Qwen2.5-VL-32B-Instruct}\\[0.5em]
            },
        overlay unbroken={
                \draw[decorate, decoration={random steps, segment length=20pt, amplitude=1pt}, line width=0.75pt]
                (frame.south west) -- (frame.north west) --
                (frame.north east) -- (frame.south east) -- cycle;
            },
        overlay first={
                \draw[decorate, decoration={random steps, segment length=20pt, amplitude=1pt}, line width=0.75pt]
                (frame.south west) -- (frame.north west) --
                (frame.north east) -- (frame.south east) -- cycle;
            },
        overlay middle={
                \draw[decorate, decoration={random steps, segment length=20pt, amplitude=1pt}, line width=0.75pt]
                (frame.south west) -- (frame.north west) --
                (frame.north east) -- (frame.south east) -- cycle;
            },
        overlay last={
                \draw[decorate, decoration={random steps, segment length=20pt, amplitude=1pt}, line width=0.75pt]
                (frame.south west) -- (frame.north west) --
                (frame.north east) -- (frame.south east) -- cycle;
            },
    ]
    \dots

    \textbf{2. Analyzing Each Option:}

    \dots

    \textbf{Option C: The figure contains 8 types of nucleotides, 5 types of bases, and 2 types of base pairings}

    \begin{itemize}[leftmargin=2em, itemsep=0.5pt, topsep=0.5pt]
        \item In the context of RNA and DNA, there are 4 types of nucleotides in each (A, U, G, C for RNA; A, T, G, C for DNA). When they anneal, the possible base pairings are A-T (or A-U), and G-C. This means there are 4 types of nucleotides per molecule, 5 types of bases (A, T, U, G, C), and 2 types of base pairings (A-T/U, G-C).
        \item \textcolor{red!75!black!}{The statement that there are 8 types of nucleotides is unreasonable because there are only 4 types of nucleotides in each (DNA and RNA).}
    \end{itemize}

    \textbf{Option D: Since segments 2, 4, and 6 cannot bind with the messenger RNA, segments 2, 4, and 6 belong to non-gene segments of the DNA molecule}

    \begin{itemize}[leftmargin=2em, itemsep=0.5pt, topsep=0.5pt]
        \item This is reasonable because segments 2, 4, and 6 do not anneal with the mRNA, which suggests that these segments do not code for the protein and are likely non-coding (non-gene) segments of the DNA.
    \end{itemize}

    \[
        \boxed{C}
    \]
\end{tcolorbox}

\begin{tcolorbox}[
        enhanced,
        sharp corners,
        colback=white,
        frame hidden,
        boxrule=0pt,
        skin=enhancedmiddle,
        before upper={
                \raisebox{-0.4em}{\includegraphics[width=1.5em]{mm-eureka-logo.png}}~\textbf{MM-EUREKA-32B}\\[0.5em]
            },
        overlay unbroken={
                \draw[decorate, decoration={random steps, segment length=20pt, amplitude=1pt}, line width=0.75pt]
                (frame.south west) -- (frame.north west) --
                (frame.north east) -- (frame.south east) -- cycle;
            },
        overlay first={
                \draw[decorate, decoration={random steps, segment length=20pt, amplitude=1pt}, line width=0.75pt]
                (frame.south west) -- (frame.north west) --
                (frame.north east) -- (frame.south east) -- cycle;
            },
        overlay middle={
                \draw[decorate, decoration={random steps, segment length=20pt, amplitude=1pt}, line width=0.75pt]
                (frame.south west) -- (frame.north west) --
                (frame.north east) -- (frame.south east) -- cycle;
            },
        overlay last={
                \draw[decorate, decoration={random steps, segment length=20pt, amplitude=1pt}, line width=0.75pt]
                (frame.south west) -- (frame.north west) --
                (frame.north east) -- (frame.south east) -- cycle;
            },
    ]
    To solve this problem, we need to analyze the image and combine it with the principle of hybridization between DNA and messenger RNA (mRNA). Let's reason step by step:

    \dots

    \textbf{2. Analyzing the options}:

    \dots
    
    \textbf{C. The image contains 8 types of nucleotides, 5 types of bases, and 2 types of base pairings}:
    \begin{itemize}[leftmargin=2em, itemsep=0.5pt, topsep=0.5pt]
        \item Nucleotides:
              \begin{itemize}[leftmargin=2em, itemsep=0.5pt, topsep=0.5pt]
                  \item \textcolor{green!45!black!}{DNA contains 4 types of deoxyribonucleotides (dA, dT, dC, dG)}
                  \item \textcolor{green!45!black!}{mRNA contains 4 types of ribonucleotides (A, U, C, G)}
                  \item \textcolor{green!45!black!}{Total: 8 types of nucleotides.}
              \end{itemize}
        \item Bases:
              \begin{itemize}[leftmargin=2em, itemsep=0.5pt, topsep=0.5pt]
                  \item DNA contains 4 bases (A, T, C, G)
                  \item mRNA contains 4 bases (A, U, C, G)
                  \item Since U replaces T in RNA, there are only 5 distinct bases (A, T/U, C, G).
              \end{itemize}
        \item Base pairing:
              \begin{itemize}[leftmargin=2em, itemsep=0.5pt, topsep=0.5pt]
                  \item In DNA-mRNA hybridization, the pairings are A–U, T–A, C–G, G–C.
                  \item However, only A–U and C–G actually form during hybridization.
                  \item Thus, only 2 types of base pairings exist.
              \end{itemize}
    \end{itemize}

    \textbf{D. Since fragments 2, 4, and 6 do not hybridize with mRNA, they belong to non-genic segments of DNA}:
    \begin{itemize}[leftmargin=2em, itemsep=0.5pt, topsep=0.5pt]
        \item The unhybridized regions (2, 4, 6) do not pair with mRNA, but this does not directly imply that they are non-genic segments.
        \item In genomic DNA, unpaired regions could be non-coding parts of a gene (e.g., introns, enhancers), or regulatory regions.
        \item Therefore, we cannot conclude that they are non-genic simply because they do not hybridize.
    \end{itemize}

    \textbf{3. Conclusion}:
    \begin{itemize}[leftmargin=2em, itemsep=0.5pt, topsep=0.5pt]
        \item Option D is incorrect because we cannot determine from hybridization alone that these regions are non-genic.
    \end{itemize}

    \textless answer\textgreater The answer is $ \boxed{D} $. \textless /answer\textgreater
\end{tcolorbox}}
    \caption{Example from biology. \textcolor{red!75!black!}{Red} highlights Qwen2.5-VL-32B-Instruct's reasoning errors. \textcolor{green!45!black!}{Green} shows MM-Eureka-32B's correct reasoning.}
    \label{fig:qualitative_bio}
\end{figure}

\end{document}